\documentclass[runningheads]{llncs}
\usepackage{graphicx}
\usepackage[table,usenames,dvipsnames]{xcolor}
\usepackage{tikz}
\usepackage{comment}
\usepackage{amsmath,amssymb} % define this before the line numbering.
\usepackage{color}
\usepackage{amsmath,graphicx}
\usepackage{tabularx,ragged2e}
\usepackage{csquotes}
\usepackage{arydshln}
\usepackage{caption}

\usepackage{cite}
\usepackage{subfiles}
\usepackage{fontawesome}
\usepackage{multirow}
\usepackage{inconsolata}
\usepackage{epsfig}
\usepackage{amsmath}
\usepackage{amssymb}
\usepackage{multirow}
\usepackage{booktabs}
\usepackage{cellspace}
\usepackage{subcaption}
\usepackage{array}
\usepackage[hang,flushmargin]{footmisc}

\usepackage{booktabs,tabularx, colortbl} 
\usepackage{collcell}
\usepackage{pgfplots}
\pgfplotsset{compat=newest}
\usepgfplotslibrary{colormaps}
\usepackage{enumitem}
\usepackage{adjustbox}

\usepackage[accsupp]{axessibility}  % Improves PDF readability for those with disabilities.

\definecolor{heatmap-yellow}{rgb}{1.0, 1.0, 0.8509803921568627}
\definecolor{heatmap-blue}{rgb}{0.03137254901960784, 0.11372549019607843, 0.34509803921568627}

\newcommand{\etal}{\textit{et al}. }
\newcommand{\ie}{\textit{i}.\textit{e}. }
\newcommand{\eg}{\textit{e}.\textit{g}. }

\usepackage{xcolor,pifont}

\usepackage[table]{xcolor}
\definecolor{mydarkblue}{rgb}{0,0.08,0.45}
\usepackage[pagebackref,
            breaklinks,
            colorlinks,
            linkcolor=mydarkblue,
            citecolor=mydarkblue,
            filecolor=mydarkblue,
            urlcolor=mydarkblue,]{hyperref}
\usepackage[capitalise]{cleveref}

\newcommand*\colourcheck[1]{%
  \expandafter\newcommand\csname #1check\endcsname{\textcolor{#1}{\ding{51}}}%
}
\newcommand*\colourcross[1]{%
  \expandafter\newcommand\csname #1cross\endcsname{\textcolor{#1}{\ding{55}}}%
}
\colourcheck{blue}
\colourcheck{green}
\colourcross{red}

\definecolor{darkgreen}{rgb}{0.0, 0.75, 0.0}

\definecolor{bananayellow}{rgb}{1.0, 0.88, 0.21}

\usepackage{multirow}

\definecolor{lowcolor}{HTML}{ef8a62}
\definecolor{centralcolor}{HTML}{f7f7f7}
\definecolor{highcolor}{HTML}{67a9cf}
\definecolor{same_attr}{HTML}{009900}
\definecolor{diff_attr}{HTML}{FE0000}

\pgfplotsset{%
  colormap={PiYG}{%
    rgb=(0.93725490196, 0.54117647058, 0.38431372549)%
    rgb=(0.96862745098, 0.96862745098, 0.96862745098)%
    rgb=(0.40392156862, 0.66274509803, 0.81176470588)%
  }%
}

\def\pgfplotsshowcolormap#1{%
    \pgfplotscolormapifdefined{#1}{\relax}{%
        \pgfplotsset{colormap/#1}%
    }%
    \pgfplotscolormaptoshadingspec{#1}{1cm}\result
    \def\tempb{\pgfdeclarehorizontalshading{tempshading}{0.2cm}}%
    \expandafter\tempb\expandafter{\result}%
    \pgfuseshading{tempshading}%
}

\newlength{\savedtabcolsep}
\newcommand{\midsepremove}{\aboverulesep=0mm \belowrulesep=0mm}%
\usepackage{listofitems}
\setsepchar{ }

\newcommand{\ApplyGradient}[1]{%
    \readlist*\mylist{#1}%
  \pgfmathsetmacro{\Val}{\mylist[1]}
  \pgfmathsetmacro{\MiddleVal}{\MinVal + (\MaxVal - \MinVal) / 2}%
  \ifdim \Val pt > \MiddleVal pt%
      \pgfmathsetmacro{\PercentColor}{max(min(100.0*(\Val - \MiddleVal)/(\MaxVal-\MiddleVal),100.0),0.00)}%
      \edef\HeatCell{\noexpand\cellcolor{highcolor!\PercentColor!centralcolor}}%
      \HeatCell$#1$%
  \else
      \pgfmathsetmacro{\PercentColor}{max(min(100.0*(\MiddleVal - \Val)/(\MiddleVal-\MinVal),100.0),0.00)}%
      \edef\HeatCell{\noexpand\cellcolor{lowcolor!\PercentColor!centralcolor}}%
      \HeatCell$#1$%
  \fi%
}

\newcommand{\ApplyGradientReverse}[1]{%
    \readlist*\mylist{#1}%
  \pgfmathsetmacro{\Val}{\mylist[1]}%
  \pgfmathsetmacro{\MiddleVal}{\MinVal + (\MaxVal - \MinVal) / 2}%
  \ifdim \Val pt > \MiddleVal pt%
      \pgfmathsetmacro{\PercentColor}{max(min(100.0*(\Val - \MiddleVal)/(\MaxVal-\MiddleVal),100.0),0.00)}%
      \edef\HeatCell{\noexpand\cellcolor{lowcolor!\PercentColor!centralcolor}}%
      \HeatCell$#1$%
  \else
      \pgfmathsetmacro{\PercentColor}{max(min(100.0*(\MiddleVal - \Val)/(\MiddleVal-\MinVal),100.0),0.00)} %
      \edef\HeatCell{\noexpand\cellcolor{highcolor!\PercentColor!centralcolor}}%
      \HeatCell$#1$%
  \fi%
}

\newcolumntype{\C}[2]{>{\def\MinVal{#1}\def\MaxVal{#2}\collectcell\ApplyGradient}c<{\endcollectcell}}

\newcolumntype{\CR}[2]{>{\def\MinVal{#1}\def\MaxVal{#2}\collectcell\ApplyGradientReverse}c<{\endcollectcell}}

\usepackage[many]{tcolorbox}
\newtcolorbox{myboxi}[2][]{
  breakable,
  title=#1,
  colback=white,
  colbacktitle=white,
  coltitle=black,
  fonttitle=\itshape,
  bottomrule=-0.1pt,
  toprule=-0.2pt,
  leftrule=5pt,
  rightrule=0pt,
  titlerule=0pt,
  arc=0pt,
  boxsep=1mm,
  outer arc=0pt,
  colframe=#2,
}

\begin{document}
% \renewcommand\thelinenumber{\color[rgb]{0.2,0.5,0.8}\normalfont\sffamily\scriptsize\arabic{linenumber}\color[rgb]{0,0,0}}
% \renewcommand\makeLineNumber {\hss\thelinenumber\ \hspace{6mm} \rlap{\hskip\textwidth\ \hspace{6.5mm}\thelinenumber}}
% \linenumbers
\pagestyle{headings}
\mainmatter
\def\ECCVSubNumber{1980}  % Insert your submission number here

\title{How Severe is Benchmark-Sensitivity in Video Self-Supervised Learning?}

%Benchmarking Self-supervised Video Representation Learning} % Replace with your title

% INITIAL SUBMISSION 
%\begin{comment}
%\titlerunning{ECCV-22 submission ID \ECCVSubNumber} 
%\authorrunning{ECCV-22 submission ID \ECCVSubNumber} 
%\author{Anonymous ECCV submission}
\author{Fida Mohammad Thoker, Hazel Doughty, Piyush Bagad, Cees Snoek}
%\institute{Paper ID \ECCVSubNumber}
\institute{University of Amsterdam}
\authorrunning{F.M. Thoker et al.}
%\end{comment}
%******************

% CAMERA READY SUBMISSION
\begin{comment}
\titlerunning{Abbreviated paper title}
% If the paper title is too long for the running head, you can set
% an abbreviated paper title here
%
\author{First Author\inst{1}\orcidID{0000-1111-2222-3333} \and
Second Author\inst{2,3}\orcidID{1111-2222-3333-4444} \and
Third Author\inst{3}\orcidID{2222--3333-4444-5555}}
%
\authorrunning{F. Author et al.}
% First names are abbreviated in the running head.
% If there are more than two authors, 'et al.' is used.
%
\institute{Princeton University, Princeton NJ 08544, USA \and
Springer Heidelberg, Tiergartenstr. 17, 69121 Heidelberg, Germany
\email{lncs@springer.com}\\
\url{http://www.springer.com/gp/computer-science/lncs} \and
ABC Institute, Rupert-Karls-University Heidelberg, Heidelberg, Germany\\
\email{\{abc,lncs\}@uni-heidelberg.de}}
\end{comment}
%******************
\maketitle

% Commented our for Appendix submission
\begin{abstract}
Despite the recent success of video self-supervised learning models, there is much still to be understood about their generalization capability. In this paper, we investigate how sensitive video self-supervised learning is to the current conventional benchmark and whether methods generalize beyond the canonical evaluation setting. We do this across four different factors of sensitivity: domain, samples, actions and task. Our study which encompasses over 500 experiments on 7 video datasets, 9 self-supervised methods and 6 video understanding tasks, reveals that current benchmarks in video self-supervised learning are not good indicators of generalization along these sensitivity factors. Further, we find that self-supervised methods considerably lag behind vanilla supervised pre-training, especially when domain shift is large and the amount of available downstream samples are low. From our analysis we distill the \textit{SEVERE-benchmark}, a subset of our experiments, and discuss its implication for evaluating the generalizability of representations obtained by existing and future self-supervised video learning methods. Code is available at \href{https://github.com/fmthoker/SEVERE-BENCHMARK}{https://github.com/fmthoker/SEVERE-BENCHMARK}.
\keywords{Self-supervised learning, Video representation learning, Video understanding}
\end{abstract}
\section{Introduction}
Video self-supervised learning has progressed at a tremendous pace in recent years, \eg~\cite{ctp-wang2021unsupervised,mmvssl3-Afouras20b,qian2021spatiotemporal,piergiovanni2020evolving, rspnet-chen2020RSPNet,gdt-patrick2020multimodal}, as it offers a crucial starting point from which to learn. 
This is especially important for video understanding applications, where annotating large amounts of data is extremely expensive, error-prone and sensitive to annotator bias. Hence, learning video representations through self-supervision is crucial, especially for use cases where the downstream video data is limited because of the domain, task or actions the video contains. However, the majority of current works in video self-supervised learning, \eg\cite{clip-order-xu2019self,frame-order-misra2016shuffle,avid-cma-morgado2021audio, selavi-asano2020labelling,videomoco-pan2021videomoco}, do not test beyond standard benchmarks. The standard protocol is to use unlabeled Kinetics-400~\cite{Kinetics-400-arxiv} for pre-training and then measure performance by finetuning on two action recognition datasets: UCF-101~\cite{UCF-101-arxiv} and HMDB-51~\cite{HMDB-51-ICCV}. While these benchmarks have facilitated the impressive progress of video self-supervised learning in recent years, they cannot indicate the generalizability of such methods as these pre-training and downstream datasets are all similar in appearance and the type of actions they contain. Some methods have started to report finetuning performance on additional datasets like Something-Something-v2~\cite{SS-v2-arxiv} in~\cite{ctp-wang2021unsupervised,rspnet-chen2020RSPNet,large-scale-feichtenhofer2021large}, Diving-48~\cite{diving} in~\cite{dave2021tclr,wang2021removing}, AVA~\cite{AVA-Gu_2018_CVPR} in~\cite{xiao2021modist,yang2020video,large-scale-feichtenhofer2021large} and EPIC-Kitchens-100~\cite{EPIC-100-arxiv} in~\cite{yang2020video}.  However, such evaluations are insufficient to understand the generalization of video self-supervised methods alone since they only add a single additional dataset, often without comparison to prior methods.

In this work, we address the essential need to gauge the sensitivity of existing video self-supervised methods to the current benchmark by thoroughly evaluating their performance for generalization across diverse downstream settings. Similar benchmarking studies have been performed for self-supervised pre-training in images~\cite{when_contrast_work,trasnferability,contrasting_contrastive,edinburgh-ericsson2021well, goyal2019scaling, yang2020transfer, kolesnikov2019revisiting, zhai2019large, asano2019critical, newell2020useful, sariyildiz2021concept, van2021benchmarking, ericsson2021self}, which investigate  model transferability~\cite{trasnferability,edinburgh-ericsson2021well,newell2020useful,image-eval5-wallace2020extending} or the importance of factors like pre-training dataset~\cite{when_contrast_work,contrasting_contrastive,goyal2019scaling} and backbone architecture~\cite{kolesnikov2019revisiting}. 
Unfortunately, lessons from these works do not directly transfer to video self-supervised learning. First, video self-supervised tasks are distinct from those of images as they are designed to understand the temporal dimension of video~\cite{rspnet-chen2020RSPNet,dave2021tclr,ctp-wang2021unsupervised,yang2020video} in addition to the spatial understanding needed in images~\cite{simclr-pmlr-v119-chen20j}. Second, video is multi-modal and several methods~\cite{gdt-patrick2020multimodal,selavi-asano2020labelling,avid-cma-morgado2021audio} are designed to exploit cross or multi-modal understanding, which is again absent in image-based methods. For videos,~\cite{large-scale-feichtenhofer2021large} extends four image-based self-supervised methods to videos and investigate their performance focusing on different pre-training setups.
We take inspiration from this and benchmarking works in image self-supervised learning and perform a much-needed study for understanding the generalizability of self-supervised methods for video in relation to different downstream factors. 

As our first contribution, we identify the problem of benchmark-sensitivity in video self-supervised learning and examine this sensitivity along the factors of domain, samples, actions and task. As our second contribution, we perform an extensive evaluation which spans a total of over 500 experiments with 9 video self-supervised learning methods across 7 video datasets and 6 video understanding tasks. We find that standard benchmarks in video self-supervised learning  do not indicate generalization along the said sensitivity factors and vanilla supervised pre-training outperforms self-supervised pre-training, particularly when domain change is large and there are only a few downstream finetuning samples available. Third, we propose a subset of our experiments as the SEVERE-benchmark for future self-supervised learning methods to benchmark generalization capability. We also discuss the implication of this benchmark for evaluating the generalizability of representations obtained by existing methods as well as the nature of video self-supervised objectives that currently generalize well. 

\begin{figure}[t]
    \centering
    \captionsetup{font=small,skip=1mm}
    \includegraphics[width=\linewidth]{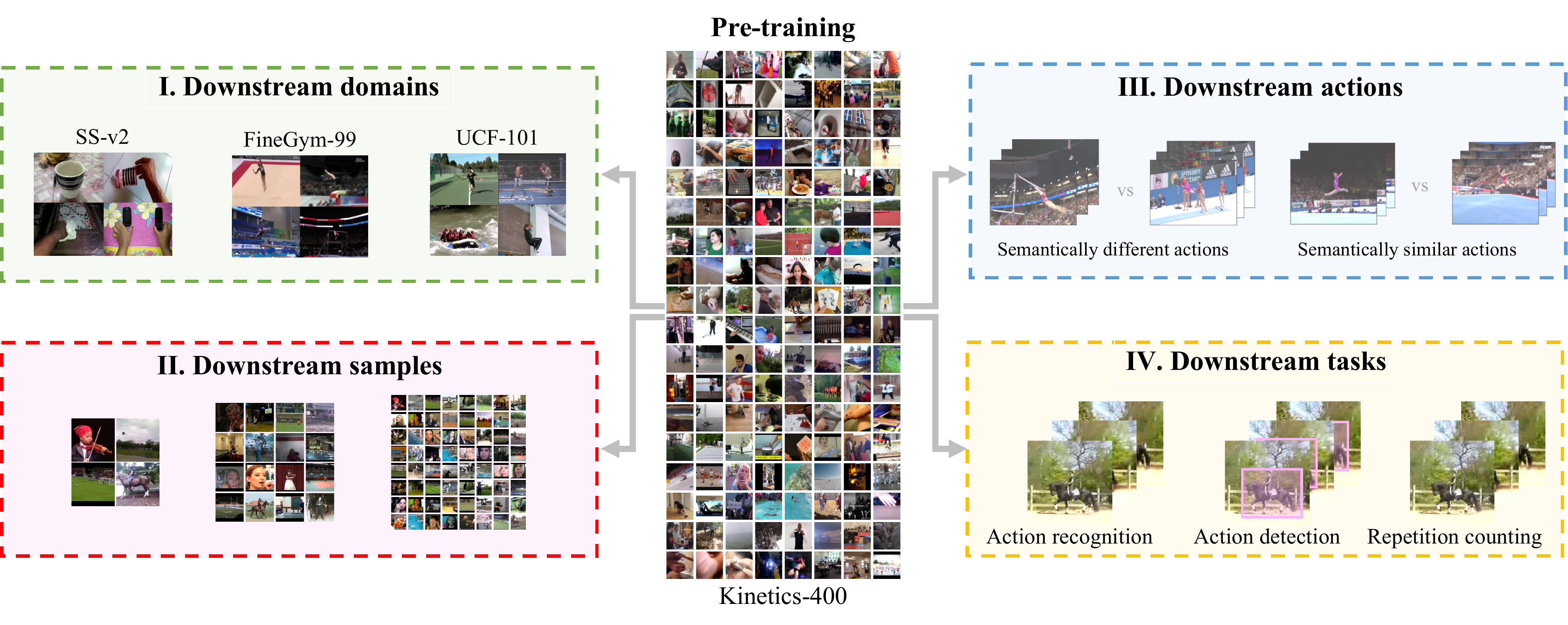}
    \caption{\textbf{Benchmark-sensitivity.} We evaluate the sensitivity of 9 video self-supervised learning methods along 4 downstream factors which vary from the pre-training source: the domain, the samples, the actions and the task.}
    \label{fig:concept-figure}
\end{figure}
\section{Identifying Benchmark Sensitivity}
The vast majority of current works in video self-supervised learning evaluate their approach by pre-training on Kinetics-400~\cite{Kinetics-400-arxiv} and finetuning the learned representation for action recognition on UCF-101\cite{UCF-101-arxiv} and HMDB-51\cite{HMDB-51-ICCV}. Some works~\cite{gdt-patrick2020multimodal, dave2021tclr, pretext-contrast-DBLP:journals/corr/abs-2010-15464, ctp-wang2021unsupervised, rspnet-chen2020RSPNet, selavi-asano2020labelling,gavrilyuk2021motion, lin2021self,huang2021ascnet} also report performance on video retrieval for UCF-101 and HMDB-51 and 
several recent works~\cite{qian2021spatiotemporal,yang2020video,recasens2021broaden} compare linear evaluation performance on Kinetics-400. However, these downstream datasets are very similar to each other and also share many similarities with the pre-training dataset of Kinetics-400. %Almost all v
Videos in all three datasets are collected from YouTube and are mostly recorded with a single camera containing a single well-positioned human actor. 
In terms of class labels, all datasets focus on similar, coarse-grained and mutually exclusive actions with many actions common between pre-training and downstream datasets. Besides all these data similarities, the existing evaluations also ignore a major benefit of self-supervised representation learning for videos, \ie finetuning the representation with only a small amount of data samples and transferring to other video understanding tasks beyond action recognition. Hence, we believe the current benchmark standard is insufficiently equipped to gain a true understanding of where video self-supervised models are successful, as it cannot show the generalizability or the sensitivity of methods to factors such as domain shift, amount of finetuning data samples, action similarity or task shift. %, etc. 
In this study, we identify the sensitivity of existing evaluations and thoroughly benchmark self-supervised video learning methods along four sensitivity factors as depicted in \cref{fig:concept-figure}.
%\medskip
\begin{enumerate}[label=\Roman*.]
\item \textbf{Downstream domain.}
First, we analyse whether features learned by self-supervised models transfer to datasets that vary in domain with respect to the pre-training dataset. 
\item \textbf{Downstream samples.} Second, we evaluate the sensitivity of self-supervised methods to the number of downstream samples available for finetuning. 
\item \textbf{Downstream actions.}
Third, we investigate if self-supervised methods learn fine-grained features required to recognize semantically similar actions. 
\item \textbf{Downstream task.}
Finally, we study the sensitivity of video self-supervised methods to the downstream task and question whether self-supervised features can be used beyond action recognition. 
\end{enumerate}

\begin{figure}[t!]
    \centering
    \includegraphics[width=\linewidth]{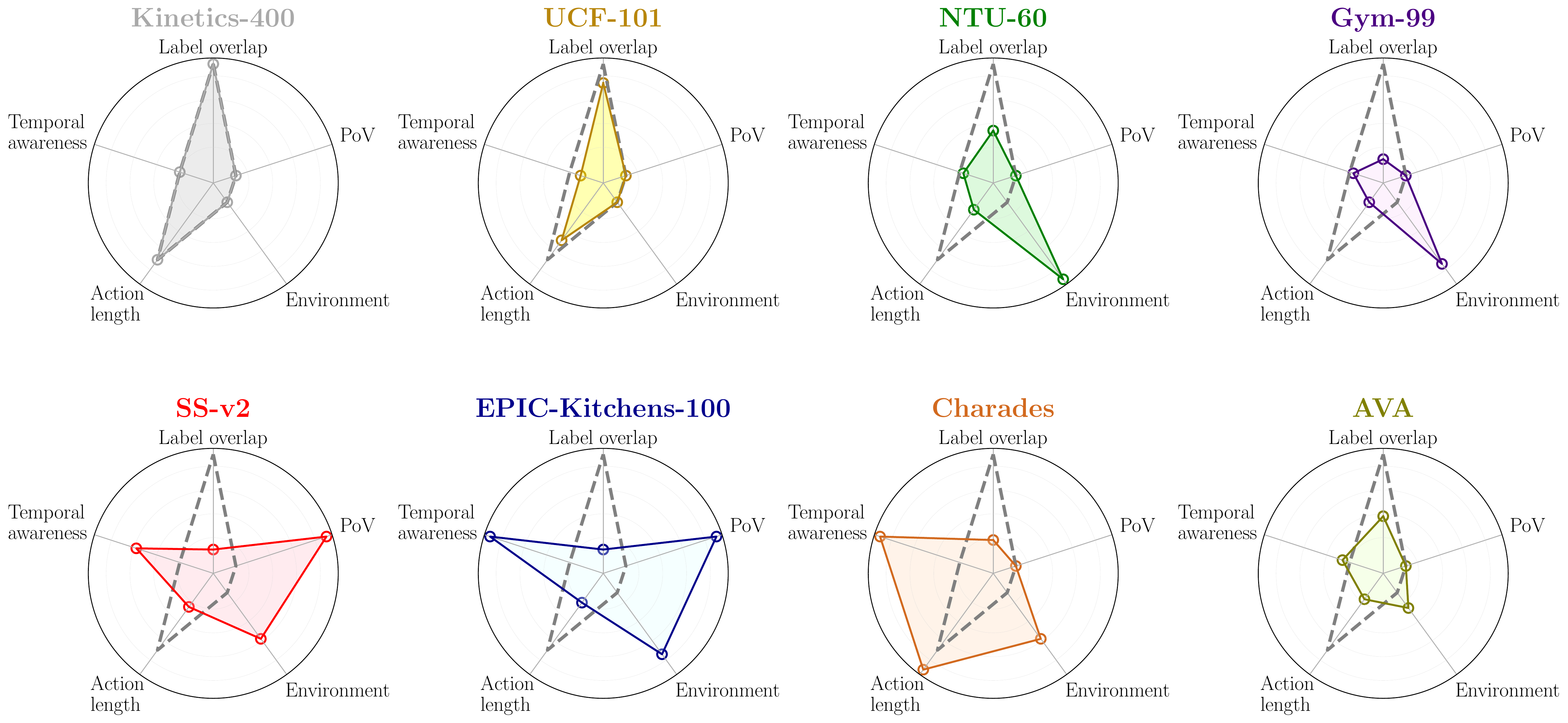}
    \caption{\small \textbf{Video dataset characteristics.}
    % \hd{Without axes labels its easy to misinterpret label overlap as the further from the center the more label overlap. I suggest reversing this one.}
    Characterizing domain shift in datasets via difference in label overlap, point-of-view (PoV), environment, action length and temporal awareness with Kinetics-400 (shown by dotted line). Kinetics-400 and UCF-101 are highly similar to each other, while datasets like Something-Something-v2, EPIC-Kitchens-100 and Charades have different attributes compared to Kinetics-400.}
    \label{fig:radar}
\end{figure}

\subsection{Downstream Video Datasets}
\label{subsec:domain-shift}
We evaluate various self-supervised models along our four sensitivity factors on 7 video datasets: \textbf{UCF-101}~\cite{UCF-101-arxiv}, \textbf{NTU-60}~\cite{NTU-60-arxiv}, \textbf{FineGym} (Gym-99) \cite{Gym-99-arxiv}, \textbf{SomethingSomething-v2} (SS-v2)~\cite{SS-v2-arxiv}, \textbf{EPIC-Kitchens-100} (EK-100)~\cite{EPIC-100-arxiv}, \textbf{Charades}~\cite{charades-sigurdsson:hal-01418216} and \textbf{AVA}~\cite{AVA-Gu_2018_CVPR}. They include a considerable variety in video domain, the actions they contain and cover a range of video understanding tasks. To get a sense of the differences between these downstream datasets and the Kinetics-400 source dataset, we summarize their similarity to Kinetics-400 by radar plots in \cref{fig:radar} based on several attributes. 
    \textit{Environment} refers to the variety of settings contained in the dataset. 
    \textit{Point-of-view} 
    is whether a video is recorded from a first-person or third-person viewpoint.
    \textit{Temporal awareness} defines the extent to which temporal context is required to recognize or detect actions. We quantify this as the point at which performance saturates with increasing temporal context in the input.
    \textit{Label overlap} is the fraction of actions in a target dataset that are also present in 
    Kinetics-400.
    \textit{Action length} is the temporal length of the actions in seconds. Details  are provided in the appendix.

\subsection{Evaluated Self-Supervised Video Learning Methods}
Self-supervised learning methods in video can be grouped into two categories based on the objective they use: pretext task methods and contrastive learning methods. Pretext task methods use predictive tasks  
such as solving spatio-temporal jigsaw puzzles~\cite{jigaw1-ahsan2019video,jigsaw2-huo2021selfsupervised, jigsaw3-kim2019self}, rotation prediction~\cite{rotate1-jing2019selfsupervised}, 
frame and clip order~\cite{frame-order-misra2016shuffle, shuffle1-fernando2017self, shuffle2-suzuki2018learning, clip-order-xu2019self,yao2021seco}, video speed~\cite{relative-speed1-benaim2020speednet, relative-speed2-jenni2020video, playback1-cho2020self, playback2-yao2020video, playback3-wang2020self}, video  completion~\cite{vcp}, predicting motion statistics~\cite{wang2019self},  tracking random patches in video frames~\cite{ctp-wang2021unsupervised} or audio-visual clustering~\cite{multimodal-clustering1-chen2021multimodal, multimodal-clustering2-hu2019deep,  selavi-asano2020labelling,alwassel2020self}.
Contrastive learning methods %~\cite{qian2021spatiotemporal, videomoco-pan2021videomoco, dave2021tclr, avid-cma-morgado2021audio, gdt-patrick2020multimodal, coclr} 
discriminate between `positive' and `negative' pairs to learn invariances to certain data augmentations and instances either from visual-only input~\cite{videomoco-pan2021videomoco,dave2021tclr,han2019video, yang2020video,qian2021spatiotemporal,lin2021self,diba2021vi2clr, sun2021composable} or multi-modal data~\cite{gdt-patrick2020multimodal,avid-cma-morgado2021audio,coclr, tao2020self,ma2021active,korbar2018cooperative,fmthoker_acmmm}.

Some methods also combine pretext and contrastive approaches~\cite{pretext-contrast-DBLP:journals/corr/abs-2010-15464, rspnet-chen2020RSPNet,pretext-contrast-2-zhang2021contrastive,taco-bai2020can, diba2021vi2clr, huang2021ascnet}. 
A detailed survey of video self-supervised learning methods can be found in~\cite{schiappa2022self}.
We consider %a total of
% nine
9
video-based self-supervised methods which achieve good performance on current benchmarks and cover a range of self-supervised paradigms in the video domain, including contrastive learning, pretext-tasks, their combination and cross-modal audio-video learning. 

Due to the high computational cost of training self-supervised methods, we focus on works with publicly available weights for a common R(2+1)D-18 network~\cite{tran2018closer} pre-trained on Kinetics-400~\cite{Kinetics-400-arxiv}:
 \textbf{MoCo}~\cite{moco_v2},
 \textbf{SeLaVi}~\cite{selavi-asano2020labelling},
 \textbf{VideoMoCo}~\cite{videomoco-pan2021videomoco},
 \textbf{Pretext-Contrast}~\cite{pretext-contrast-DBLP:journals/corr/abs-2010-15464},
 \textbf{RSPNet}~\cite{rspnet-chen2020RSPNet},
 \textbf{AVID-CMA}~\cite{avid-cma-morgado2021audio}, 
 \textbf{CtP}~\cite{ctp-wang2021unsupervised},
 \textbf{TCLR}~\cite{dave2021tclr} and
 \textbf{GDT}~\cite{gdt-patrick2020multimodal}. We compare these to no pre-training, \ie training from scratch, and fully supervised pre-training for action recognition. 
It is worth noting that since we use publicly available models we cannot control the exact pre-training setup. There are subtle differences in the training regime for each method, such as the number of epochs, the data augmentations used and the batch size. Details of these differences are provided in the appendix. However, all models use the same backbone and pre-training dataset thus we can evaluate their downstream abilities in exactly the same way. To finetune for downstream tasks we simply attach a task-dependent head at the last layer of the pre-trained R(2+1)D-18 backbone 
 to produce label predictions for the corresponding task. 
For a fair comparison, we use the same set of hyper-parameters, optimization and pre-processing during the downstream training of each model.

\section{Sensitivity Factor I: Downstream Domain}
\label{sec:factor_1}
We first investigate to what extent self-supervised methods learn features that are applicable to action recognition in any domain. 
We evaluate the suite of pre-trained models on  UCF-101, NTU-60, Gym-99, SS-v2 and EK-100  for the task of action recognition.
It is worth noting that as well as variety in domain, these datasets include variety in the amount of training data (9.5k - 168k examples) and cardinality of classification (60 - 300 classes). We attach a single classification layer to the pre-trained backbone and evaluate the models' performance on the downstream task in two settings. First, \textbf{full finetuning} where we train the whole network from the initialization of the pre-trained weights. Second, \textbf{linear evaluation} where we train the classification layer only using the frozen features of pre-trained backbones. 
We follow the standard splits proposed in the original datasets and report video-level top-1 accuracy on the test sets. The details about splits, pre-processing, training  for each dataset are provided in the appendix. 

\medskip
\noindent\textbf{Full finetuning.} The left part of \cref{domain_shift} shows the results of full finetuning. 
From the results, it is clear that all self-supervised methods are very effective on UCF-101 as there is a significant gap between training from scratch and using self-supervised pre-training. This gap is reduced as the difference between Kinetics-400 and the downstream domain increases. SeLaVi, MoCo and AVID-CMA in particular are evidence of this as these methods suffer when datasets have higher temporal awareness and less label overlap with Kinetics-400. When moving from UCF-101 to NTU-60 and Gym-99 there is a change in the ordering of self-supervised methods. This demonstrates a high performance on UCF-101 does not guarantee a self-supervised model is generalizable to other domains. The change in ranking is even more prominent for SS-v2 and EK-100, which require the most temporal awareness and also shift to a first-person viewpoint. This is particularly noticeable for AVID-CMA. On these datasets, MoCo has similar results to no pre-training, which is evidence that video-specific self-supervised learning methods are needed and that image-based methods are insufficient.
Overall, supervised pre-training achieves good performance across the board, outperforming self-supervised methods on the most similar domain (UCF-101) as well as the most dissimilar domains (SS-v2 and EK-100). Amidst the models tested, CtP, RSPNet, VideoMoCo and TCLR stand out as the self-supervised pre-training methods most generalizable to different domains.

\begin{table}[t]
\captionsetup{font=small,skip=2mm}
         \caption[]{\textbf{Sensitivity Factor I: Downstream Domain.} Video self-supervised methods evaluated across datasets with increasing domain shift with respect to the source dataset (see \cref{fig:radar}).  
         Colors denote relative rankings across methods for each dataset, ranging from \textcolor{lowcolor}{low} \begin{tikzpicture}%
      \pgfplotscolorbardrawstandalone[%
        colormap name=PiYG,%
        colorbar horizontal,%
        colorbar style={%
          height=0.18cm,%
          width=2cm,%
          hide axis,%
        }%
      ]%
    \end{tikzpicture} \textcolor{highcolor}{high}. The ranking of methods is domain-sensitive for both finetuning and linear classification and becomes less and less correlated with the current UCF-101 benchmark as the domain shift increases.}
    \centering
    \midsepremove
    \resizebox{\linewidth}{!}{\begin{tabular}{
    l\C{77.3}{93.9}\C{92.8}{94.3}\C{89.8}{92.1}\C{52.0}{60.8}\C{25.7}{47.7}c\C{7.61}{65.87}\C{37.9}{91.7}\C{15.7}{53.9}\C{20.2}{45.1}\C{4.5}{16.6}\C{20.0}{26.6}}
    \toprule
    \addlinespace[0.1cm]
     \multirow{2}{*}{\textbf{Pre-training}} & \multicolumn{5}{Sc}{\textbf{Finetuning}} & & \multicolumn{6}{Sc}{\textbf{Linear Evaluation}} \\
    \addlinespace[0.04cm]
    \cmidrule{2-6} \cmidrule{8-13}
      \addlinespace[0.1cm]
         & \multicolumn{1}{c}{UCF101} &  \multicolumn{1}{c}{NTU60} & \multicolumn{1}{c}{Gym99} & \multicolumn{1}{c}{SSv2} & \multicolumn{1}{c}{EK 100} & & \multicolumn{1}{c}{K 400} & \multicolumn{1}{c}{UCF101} &  \multicolumn{1}{c}{NTU60} & \multicolumn{1}{c}{Gym99} & \multicolumn{1}{c}{SSv2} & \multicolumn{1}{c}{EK 100}\\
         \midrule
           \addlinespace[0.01cm]
         None                    & 77.3 & 92.9 & 89.8 & 57.1 & 25.7 & & \multicolumn{1}{c}{-}& \multicolumn{1}{c}{-} & \multicolumn{1}{c}{-} & \multicolumn{1}{c}{-} & \multicolumn{1}{c}{-} & \multicolumn{1}{c}{-}\\
           \addlinespace[0.01cm]
         \midrule
        \addlinespace[0.01cm]
         MoCo                    & 83.3 & 93.4 & 90.7 & 57.1 & 26.4 & & 34.5 & 65.4 & 16.0 & 21.2 & 7.4 & 21.4 \\
         VideoMoCo               & 84.9 & 94.1 & 90.3 & 59.0 & 43.6 && 31.0 & 66.3 & 51.6 & 41.6 & 19.5 & 25.7 \\
         SeLaVi                  & 85.2 & 92.8 & 88.9 & 56.2 & 33.8 && 24.1 & 51.2 & 15.7 & 20.2 & 4.5 & 22.4 \\
         Pretext-Contrast        & 87.7 & 93.9 & 90.5 & 56.9 & 34.3 && 22.4 & 57.2 & 17.6 & 30.0 & 10.9 & 20.0 \\
         RSPNet                  & 88.7 & 93.9 & 91.1 & 59.0 & 42.7 && 46.0 & 76.6 & 33.5 & 32.2 & 12.5 & 24.9 \\
         AVID-CMA                & 88.8 & 94.0 & 90.4 & 52.0 & 29.9 && 43.5 & 78.1 & 53.9 & 45.1 & 16.1 & 22.5 \\
         CtP                     & 90.1 & 94.3 & 92.0 & 59.6 & 42.8 && 7.6 & 37.9 & 22.6 & 30.6 & 12.2 & 20.0 \\
         TCLR                    & 90.8 & 94.1 & 91.6 & 59.8 & 36.2 && 19.9 & 63.3 & 33.5 & 33.0 & 10.8 & 21.8 \\
         GDT                     & 91.3 & 93.9 & 90.5 & 58.0 & 37.3 && 38.6 & 75.7 & 38.2 & 34.2 & 11.9 & 25.3 \\
        \addlinespace[0.01cm]
         \midrule
        \addlinespace[0.01cm]
         Supervised              & 93.9 & 93.9 & 92.1 & 60.8 & 47.7 && 65.9 & 91.7 & 45.5 & 42.7 & 16.6 & 26.6 \\
        \addlinespace[0.01cm]
         \bottomrule
    \end{tabular}
    }

    \label{domain_shift}
\end{table}

\medskip
\noindent\textbf{Linear classification.} The right part of \cref{domain_shift} shows the results for linear classification.
As with finetuning, the ranking among the self-supervised methods changes as the domain difference between the pre-training and the downstream dataset increases. For example, VideoMoCo ranks lower than GDT and RSPNet for UCF-101 and Kinetics-400 but ranks higher than both for all other datasets. This again demonstrates that performance on UCF-101 does not give a complete picture of a self-supervised model's success. We also observe that linear evaluation on Kinetics-400, as some papers report~\cite{qian2021spatiotemporal, recasens2021broaden, yang2020video}, has the same issue since it is highly correlated to UCF-101 performance.
For UCF-101 and Kinetics-400, self-supervised models with contrastive objectives learn highly discriminative features compared to the non-contrastive models. This can be seen by comparing contrastive models AVID-CMA, GDT and RSPNet to non-contrastive SeLaVi and CtP. %
From the NTU-60 and Gym-99 results we observe that as the label overlap between the pre-training and the downstream dataset decreases, the performance gap between finetuning and linear evaluation increases considerably. 
This is true for both supervised and self-supervised pre-training.
The most generalizable methods in the linear classification setting are contrastive methods VideoMoCo and AVID-CMA as well as supervised pre-training. Interestingly, there are cases where VideoMoCo and AVID-CMA even outperform supervised pre-training, namely for NTU-60, Gym-99 and SS-v2. 

\begin{myboxi}[]{LimeGreen!30}
\paragraph{Conclusion.} We observe from \cref{domain_shift} that performance for both UCF-101 finetuning and Kinetics-400 linear evaluation is not indicative of how well a self-supervised video model generalizes to different downstream domains, with the ranking of methods changing substantially across datasets and whether full finetuning or linear classification is used.
\end{myboxi}

\section{Sensitivity Factor II: Downstream Samples}
\label{sec:factor_2}
\begin{figure}[t!]
\captionsetup{font=small,skip=2mm}
    \centering
    \includegraphics[width=\linewidth]{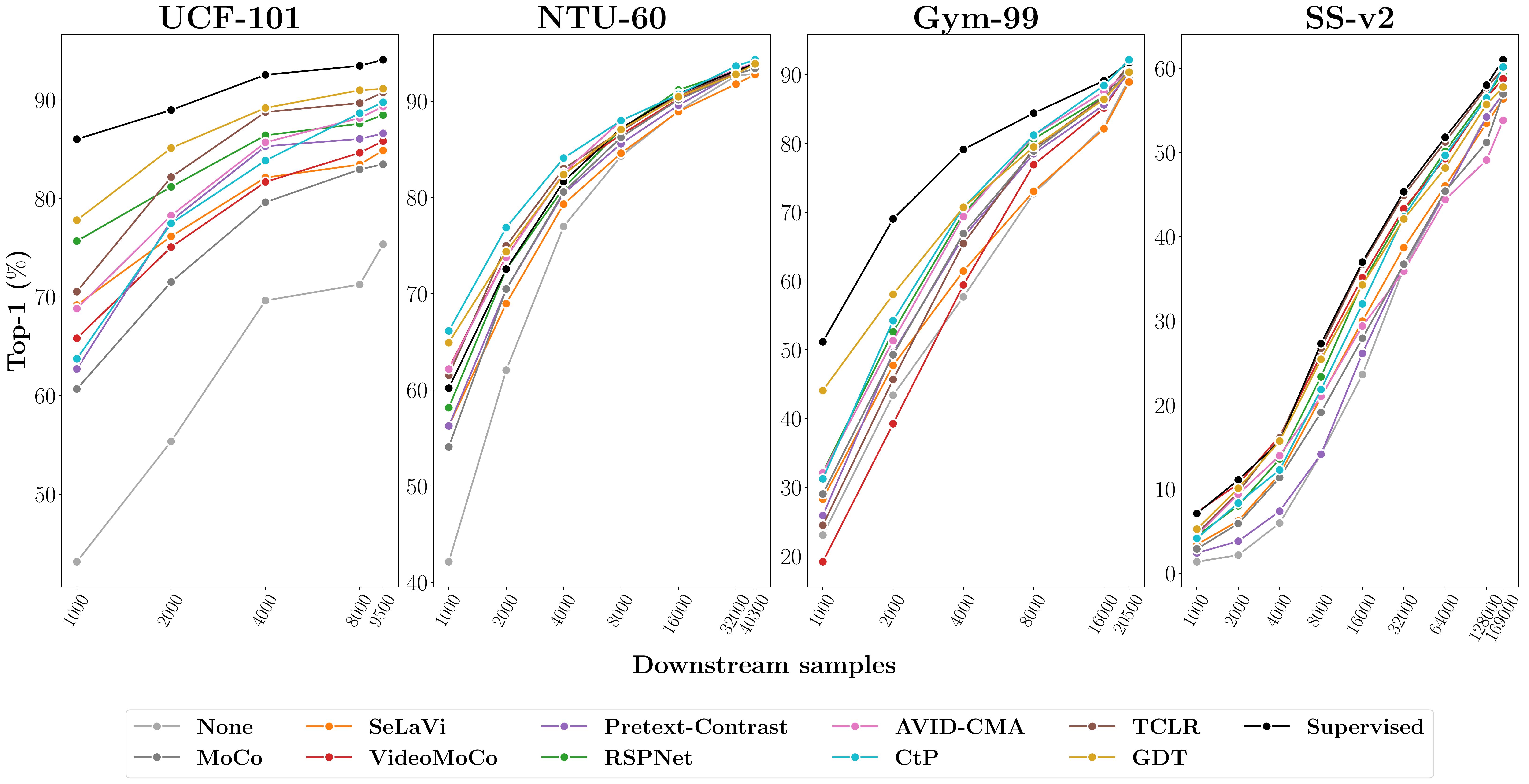}
    \caption{\textbf{Sensitivity Factor II: Downstream Samples.} Comparison of video self-supervised learning methods using varying number of finetuning samples for four downstream datasets. Both the gap and rank among pre-training methods are sensitive to the number of samples available for finetuning.
    }
    \label{fig:training-data-size}
\end{figure}

The previous section analyzed sensitivity to the downstream domain by evaluating performance on several different datasets. However, finetuning on each of these datasets uses a large number of labeled examples, which means training from scratch already obtains good performance. Not all domains and use cases have ample labeled video examples available, thus we investigate what the impact of the number of finetuning samples is and whether self-supervised methods can be beneficial in scenarios where we have little data to finetune with. We vary the amount of finetuning data, beginning from 1000 videos, sampled uniformly from the classes, and double the amount until we reach the full training set size. We report on four of the downstream datasets from the previous section: UCF-101, NTU-60, Gym-99 and SS-v2. The results are summarized in \cref{fig:training-data-size}.

We first observe that the trends in the low data regime are different from those with the full data. The gap between supervised and self-supervised pre-training is much larger in low data settings, particularly for UCF-101 and Gym-99. NTU is an exception, where, with 1000-4000 samples CtP, GDT, AVID-CMA and TCLR outperform supervised pre-training. 
As with changes in the downstream domain, change in the amount of downstream examples also causes a change in the ranking of self-supervised models. For example, on UCF-101, RSPNet is much more successful than CtP and TCLR  when using only 1000 samples. 
 This is because some self-supervised models benefit more than others from an increased amount of downstream samples. For example, CtP is one of the most generalizable pre-training strategies when finetuning with the full data on UCF-101, Gym-99 and SS-v2, but this is not the case with fewer training samples. 
Interestingly, GDT is consistently high in the ranking with low amounts of finetuning samples. This is likely due to the large number of temporal augmentations it uses, which help the generalization ability when the training data is limited.

\begin{myboxi}[]{red!10}
\paragraph{Conclusion.}
We observe from  \cref{fig:training-data-size} that video self-supervised models are highly sensitive to the amount of samples available for finetuning, with both the gap and rank between methods changing considerably across sample sizes on each dataset. 
\end{myboxi}

\section{Sensitivity Factor III: Downstream Actions}
\label{sec:factor_3}
As indicated earlier, existing evaluations of self-supervised video learning methods have been limited to coarse-grained action recognition.
In this section, we investigate whether current self-supervised tasks are only effective for these types of benchmarks or whether they are able to learn features that are useful for differentiating more challenging and semantically similar actions.

FineGym~\cite{Gym-99-arxiv} provides us with an experimental setup to study sensitivity to this factor. The dataset contains different evaluations with varying levels of semantic similarity, namely action recognition \textit{across all events}, \textit{within an event} or \textit{within a set}.
Recognition \textit{across all events} uses the whole of Gym-99 containing actions from four gymnastic events. For recognition \textit{within an event} there are two subsets: Vault and Floor containing only actions from these two events. Recognition \textit{within a set} has two subsets namely FX-S1, containing different \textit{leaps-jumps-hops} in Floor, and UB-S1, which consists of types of \textit{circles} in Uneven Bars. We also experiment with the long-tailed version of FineGym, Gym-288, which adds 189 more tail classes. Details of these subsets are in the appendix. As before, we attach a classification head to the pre-trained models and finetune the whole network with the training set of each subset. In Table~\ref{granularity} we report Top-1 accuracy (mean per-class) on the testing sets following \cite{Gym-99-arxiv}. %The results are shown in \cref{granularity}.

\begin{table}[t!]
\centering
    \midsepremove
\captionsetup{font=small,skip=2mm}
\caption[]{\textbf{Sensitivity Factor III: Downstream Actions.} Video self-supervised models evaluated on different semantic similarities of action in FineGym: across events, within an event and within a set. Colors denote relative rankings across methods for each dataset, ranging from \textcolor{lowcolor}{low} \begin{tikzpicture}%
      \pgfplotscolorbardrawstandalone[%
        colormap name=PiYG,%
        colorbar horizontal,%
        colorbar style={%
          height=0.18cm,%
          width=2cm,%
          hide axis,%
        }%
      ]%
    \end{tikzpicture} \textcolor{highcolor}{high}. Many methods struggle on the within a set benchmark where actions are most semantically similar.}
\setlength{\tabcolsep}{3mm}
\resizebox{\textwidth}{!}{%
\begin{tabular}{l\C{84.5}{88.6}@{\hskip 2mm}c\C{24.7}{37.7}\C{75.9}{86.2}@{\hskip 2mm}c\C{46.6}{79.1}\C{80.9}{88.8}c\C{50.0}{58.4}}
\toprule
\addlinespace[0.04cm]
 & \multicolumn{7}{c}{\textbf{Gym99}} & & \multicolumn{1}{c}{\textbf{Gym288}} \\ 
\addlinespace[0.04cm]
\cmidrule{2-8}\cmidrule{10-10} 
\addlinespace[0.04cm]
\multicolumn{1}{l}{\textbf{Pre-training}} & \multicolumn{1}{c}{Across Events} 
& & \multicolumn{2}{c}{Within Event} && \multicolumn{2}{c}{Within Set} & &  \multicolumn{1}{c}{Across Events}\\ 
\addlinespace[0.04cm]
\cmidrule{2-2}\cmidrule{4-5}\cmidrule{7-8}\cmidrule{10-10}
\addlinespace[0.04cm]
\multicolumn{1}{c}{} & \multicolumn{1}{c}{All} && \multicolumn{1}{c}{Vault} & \multicolumn{1}{c}{Floor} && \multicolumn{1}{c}{FX-S1} & \multicolumn{1}{c}{UB-S1} &  &  \multicolumn{1}{c}{All} \\ 
\addlinespace[0.02cm]
\arrayrulecolor{black}\midrule
\addlinespace[0.01cm]
% \arrayrulecolor{lightgray}\specialrule{0.05pt}{0.15pc}{0.15pc}
None                & 84.8 && 24.7 & 75.9 && 46.6 & 82.3  & &50.0  \\
        \addlinespace[0.01cm]\midrule        \addlinespace[0.01cm]
SeLaVi              & 84.5 && 25.4 & 76.0 && 51.3 & 80.9 && 52.8 \\
AVID-CMA            & 85.7 && 30.4 & 82.7 && 68.0 & 87.3 && 52.5 \\
VideoMoCo           & 85.9 && 28.4 & 79.5 && 57.3 & 83.9 && 54.1 \\
Pretext-contrast    & 86.0 && 28.5 & 81.4 && 66.1 & 86.1 && 52.7 \\
MoCo                & 86.5 && 33.2 & 83.3 && 65.0 & 84.5 && 55.1 \\
GDT                 & 86.6 && 36.9 & 83.6 && 66.0 & 83.4 && 55.4 \\
RSPNet              & 86.9 && 33.4 & 82.7 && 65.4 & 83.6 && 55.2 \\
TCLR                & 87.7 && 29.8 & 84.3 && 60.7 & 84.7 && 55.4 \\
CtP                 & 88.1 && 26.8 & 86.2 && 79.1 & 88.8 && 56.5 \\
        \addlinespace[0.01cm]\midrule        \addlinespace[0.01cm]
Supervised          & 88.6 && 37.7 & 86.1 && 79.0 & 87.1 & & 58.4 \\
        \addlinespace[0.01cm]
\bottomrule
\end{tabular}%
}
\label{granularity}
\end{table}

Performance of self-supervised methods varies considerably across downstream actions. The methods that perform best on Gym-99 often do not generalize well to the subsets with higher semantic similarity among actions. This is particularly noticeable for RSPNet and TCLR which drop in the ranking for the within-set subsets. All self-supervised methods, except GDT, struggle on Vault, likely due to the intense motions. % of this action.
Surprisingly, MoCo 
performs reasonably well when actions are more semantically similar, and is comparable to GDT and RSPNet.  
The best self-supervised method for subsets with high semantic similarity is CtP.
This is especially evident from FX-S1 where it outperforms the second-best self-supervised method, AVID-CMA, by 12\%. As with downstream domain and samples, supervised pre-training generalizes better than self-supervised methods across downstream actions with only CtP achieving comparable performance.

\cref{granularity} also compares balanced Gym-99 with long-tailed Gym-288. We observe that self-supervised methods are not robust to this change in distribution, with the gap in performance with respect to supervised pre-training increasing. However, the ranking remains consistent, meaning the performance on the balanced set is generally indicative of the performance on the long-tailed set.

\begin{myboxi}[]{NavyBlue!15}
\paragraph{Conclusion.} Most self-supervised methods in \cref{granularity} are sensitive to the actions present in the downstream dataset and do not generalize well to more semantically similar actions. This further emphasizes the  need for proper evaluation of self-supervised methods beyond current coarse-grained action classification. 
\end{myboxi}

\section{Sensitivity Factor IV: Downstream Tasks}
\label{sec:factor_4}
The fourth factor we  
investigate is whether self-supervised video models are sensitive to the downstream task or whether features learned by self-supervised models are useful to video understanding tasks beyond action recognition. 
We evaluate this in two ways. First, we keep the domain fixed and evaluate different tasks in a domain similar to the pre-training dataset. We also explore further tasks by changing the domain and seeing how these two factors interplay.

\subsubsection{Task-shift within domain.}
We consider three different tasks which are all defined for UCF-101: spatio-temporal action detection~\cite{yowo}, repetition counting~\cite{rep_counting} and arrow-of-time prediction~\cite{arrow_of_time}. Using UCF-101 allows us to keep the domain fixed across tasks and eliminates the impact of domain shift. Note that each task uses a different subset of the full UCF-101 dataset, however, the domain remains consistent. For each task, we use the R(2+1)D-18 networks as the pre-trained backbones as before  and attach task-dependent heads. We report mean Average Precision for spatio-temporal localization~\cite{mettes2016spot}, mean absolute counting error for repetition counting~\cite{rep_counting} and classification accuracy for arrow-of-time prediction~\cite{arrow_of_time, aot2-wei2018learning}. Further details are in the appendix.

\begin{table}[t]
\centering
    \midsepremove
\captionsetup{font=small,skip=2mm}
\caption[]{\textbf{Sensitivity Factor IV: Downstream Tasks.} Transferability of self-supervised video learning methods across video understanding tasks. Colors denote relative rankings across methods for each task, ranging from \textcolor{lowcolor}{low} \begin{tikzpicture}%
      \pgfplotscolorbardrawstandalone[%
        colormap name=PiYG,%
        colorbar horizontal,%
        colorbar style={%
          height=0.18cm,%
          width=2cm,%
          hide axis,%
        }%
      ]%
    \end{tikzpicture} \textcolor{highcolor}{high}. Note that for repetition counting lower (error) is better.  
    Self-supervised features are transferable to different  downstream tasks when the domain shift is low, but struggle when there is also a domain shift. Action recognition on UCF-101 is not a good proxy for self-supervised video learning use cases where a downstream domain- and task-shift can be expected. }
\setlength{\tabcolsep}{3mm}
\resizebox{\textwidth}{!}{%
\begin{tabular}{l\C{77.3}{93.1}\C{0.327}{0.482}\CR{0.123}{0.217}\C{56.1}{87.0}c\C{7.9}{23.5}\C{7.4}{17.9}}
\toprule
\addlinespace[0.07cm]
 & \multicolumn{4}{c}{\textbf{Task-shift within domain}} & & \multicolumn{2}{c}{\textbf{Task-shift out of domain}} \\ 
\addlinespace[0.04cm]
\cmidrule{2-5}\cmidrule{7-8} 
\addlinespace[0.04cm]
\addlinespace[0.04cm]
\multicolumn{1}{l}{\textbf{Pre-training}} & \multicolumn{1}{c}{Action} & \multicolumn{1}{c}{Action} & \multicolumn{1}{c}{Repetition} &\multicolumn{1}{c}{Arrow of} & & \multicolumn{1}{c}{Multi-label}  &  \multicolumn{1}{c}{Action} \\ 
\multicolumn{1}{c}{} & \multicolumn{1}{c}{Recognition} & \multicolumn{1}{c}{Detection} & \multicolumn{1}{c}{Counting} &\multicolumn{1}{c}{Time} & & \multicolumn{1}{c}{Recognition}  &  \multicolumn{1}{c}{Detection} \\ 
\addlinespace[0.02cm]
\midrule
\addlinespace[0.01cm]
% \arrayrulecolor{lightgray}\specialrule{0.05pt}{0.15pc}{0.15pc}
None                   & 77.3 & 0.327 & 0.217 & 56.1 && 7.9  & 7.4  \\
\addlinespace[0.01cm]\midrule        \addlinespace[0.01cm]
MoCo                   & 83.3 & 0.416 & 0.208 & 80.3 && 8.3  & 11.7 \\
VideoMoCo              & 84.9 & 0.440 & 0.185 & 72.9 && 10.5 & 13.1 \\
SeLaVi                 & 85.2 & 0.419 & 0.162 & 77.4 && 8.4  & 10.2 \\
Pretext-contrast       & 87.7 & 0.462 & 0.164 & 77.2 && 8.9  & 12.7 \\
RSPNet                 & 88.7 & 0.467 & 0.145 & 87.0 && 9.0  & 14.1 \\
AVID-CMA               & 88.8 & 0.435 & 0.148 & 83.3 && 8.2  & 10.0 \\
CtP                    & 90.1 & 0.465 & 0.178 & 77.1 && 9.6  & 10.0 \\
TCLR                   & 90.8 & 0.476 & 0.142 & 85.6 && 12.2 & 10.8 \\
GDT                    & 91.3 & 0.463 & 0.123 & 76.4 && 8.5  &  12.6 \\
\addlinespace[0.01cm]\midrule         \addlinespace[0.01cm]
Supervised             & 93.9 & 0.482 & 0.132 & 77.0 && 23.5 & 17.9 \\
\addlinespace[0.01cm]
\bottomrule
\end{tabular}%
}
\label{task_shift}
\end{table}

From the results in \cref{task_shift}, we observe that self-supervised learning is beneficial to tasks beyond action recognition, with almost all methods outperforming training from scratch on spatio-temporal action detection, repetition counting and arrow-of-time prediction. Action detection results are well correlated with action recognition. Repetition counting and arrow-of-time have less correlation with action recognition, suggesting that the current benchmark on UCF-101 action recognition by itself is not a good indication of how well self-supervised methods generalize to other tasks. For repetition counting and arrow-of-time prediction, some methods perform comparably to or outperform supervised pre-training. Notably, RSPNet and TCLR generalize the best across these tasks, with GDT also performing well on repetition counting. CtP ranks high on action recognition and detection but performs modestly for repetition counting. This shows that different methods have different task sensitivity, so a thorough evaluation along downstream tasks is needed.

\subsubsection{Task-shift out of domain.} 
We also evaluate how well the self-supervised models generalize when both the domain and the task change. We do so with two popular video understanding benchmarks: long-term multi-label classification on Charades \cite{charades-sigurdsson:hal-01418216} and short-term spatio-temporal action detection on AVA \cite{AVA-Gu_2018_CVPR}. For both, we follow the setup and training procedure from  \cite{Feichtenhofer2019SlowFastNF} with R(2+1)D-18 models as the pre-trained backbone and we measure performance in mean Average Precision. Details are in the appendix. 

From the results in \cref{task_shift}, we observe that supervised pre-training is far more generalizable than all self supervised methods, which all struggle considerably when both the domain and task change. For long-term action classification on Charades, TCLR is slightly better than other methods. On AVA, RSPNet is the best performing self-supervised method with VideoMoCo second. In \cref{sec:factor_1}, we earlier observed that these were two of the methods more robust to domain shift suggesting that this factor is key to success on AVA.

\begin{myboxi}[]{Goldenrod!50}
\paragraph{Conclusion.} The results in \cref{task_shift} reveal that action classification performance on UCF-101 is mildly indicative for transferability of self-supervised features to other tasks on UCF-101. However, when methods pre-trained on Kinetics-400 are  confronted with a domain change in addition to the task change, UCF-101 results are no longer a good proxy and the gap between supervised and self-supervised pre-training is large. 
\end{myboxi}

\section{SEVERE-benchmark}
As evident from the results in previous sections, current video self-supervised methods are benchmark-sensitive to the four factors we have studied. Based on our findings, we propose the SEVERE-benchmark (\underline{SE}nsitivity of \underline{V}id\underline{E}o \underline{RE}presentations) for use in future works to more thoroughly evaluate new video self-supervised methods for generalization along the four sensitivity factors we have examined. Since we do not expect future works to run all the experiments from our study, we create a subset of experiments that are indicative benchmarks for each sensitivity factor and realistic to run. We summarize the benchmark composition in \cref{proposed-benchmarks} and detail its motivation per factor. Standard deviations for the results we obtain on this benchmark can be found in the appendix.\\
\noindent\textbf{Downstream domain.} 
To measure a self-supervised model's domain sensitivity we recommend 
using Something-Something-v2 and FineGym-99. These two datasets come from domains distinct to Kinetics-400 and UCF-101 and also each other. FineGym-99 evaluates a model's ability to generalize to datasets with less distinctive backgrounds where there are few actions in common with Kinetics-400. SS-v2 evaluates the generalizability to actions that require high temporal awareness as well as the shift to a first-person viewpoint. It is evident from \cref{proposed-benchmarks} that there are significant rank changes between  UCF-101, Gym-99 and SS-v2 thus these three datasets provide a challenging subset for future methods. \\
\noindent\textbf{Downstream samples.} 
For the sample sensitivity, we recommend using 1000 samples on UCF-101 and Gym-99. Using 1000 samples showed the most dramatic difference from the full dataset size particularly for these datasets where there is a considerable gap between self-supervised and supervised pre-training as well as considerable rank change among the methods. \\
\noindent\textbf{Downstream actions.}
To test generalizability to recognizing semantically similar actions, we recommend evaluating the two within-set granularities of Gym-99 \ie FX-S1 and UB-S1. 
Both of these subsets have high semantic similarity between actions with methods currently struggling to generalize to both of these subsets as can be seen in 
\cref{proposed-benchmarks}. 
There is also a significant gap between supervised and most self-supervised pre-training methods for FX-S1, highlighting the potential for future works in this area.\\
\noindent\textbf{Downstream task.} 
To evaluate the task sensitivity, we recommend using repetition counting on UCF-101 and multi-label classification on Charades. Repetition counting on UCF-101 highlights different strengths to action recognition as it allows investigation of a model's ability to generalize to a task that requires more temporal understanding without measuring the impact of the domain. We recommend multi-label classification on Charades as it is currently a very challenging task for self-supervised models and allows the combination of domain and task shift to be investigated.  
Code to compare on the SEVERE-benchmark is available  at \href{https://github.com/fmthoker/SEVERE-BENCHMARK}{https://github.com/fmthoker/SEVERE-BENCHMARK}.

\begin{table}[t]
\captionsetup{font=small,skip=2mm}
         \caption[]{\textbf{Proposed SEVERE-benchmark} for evaluating video self-supervised methods for generalization along downstream domains, samples, actions and tasks.
    }
    \centering
    \midsepremove
    \resizebox{\linewidth}{!}{\begin{tabular}
    {
    l\C{77.3}{93.9}c
    \C{52.0}{60.8}\C{89.9}{92.1}@{\hskip 2mm}c
    \C{38.3}{86.6}\C{22.7}{51.3}@{\hskip 2mm}c
    \C{46.6}{79.1}\C{80.9}{88.8}@{\hskip 2mm}c
    \CR{0.123}{0.217}\C{7.9}{23.5}
    }
    % {lc cc ccccccc}
    \toprule
    \addlinespace[0.1cm]
     & \multicolumn{1}{Sc}{\textbf{Existing}} & & \multicolumn{11}{Sc}{\textbf{SEVERE-benchmark}} \\
    \addlinespace[0.04cm]
    \cmidrule{2-2} \cmidrule{4-14}
      \addlinespace[0.1cm]
         %& \multicolumn{1}{c}{} & & \multicolumn{2}{Sc}{Domains} &  & \multicolumn{2}{Sc}{Samples} & & \multicolumn{2}{Sc}{Actions}& & \multicolumn{2}{Sc}{Tasks}\\
         \multicolumn{1}{l}{\textbf{Pre-training}} & \multicolumn{1}{c}{} & & \multicolumn{2}{Sc}{Domains} &  & \multicolumn{2}{Sc}{Samples} & & \multicolumn{2}{Sc}{Actions}& & \multicolumn{2}{Sc}{Tasks}\\
    %   \cmidrule{4-11}
      \cmidrule(lr){4-5} \cmidrule(lr){7-8} \cmidrule(lr){10-11} \cmidrule(lr){13-14}
      \addlinespace[0.1cm]
         & \multicolumn{1}{c}{UCF101} & & \multicolumn{1}{c}{SS-v2} & \multicolumn{1}{c}{Gym-99} & & \multicolumn{1}{c}{UCF ($10^{3}$)} & \multicolumn{1}{c}{Gym-99 ($10^{3}$)} & & \multicolumn{1}{c}{FX-S1 } & \multicolumn{1}{c}{UB-S1}& & \multicolumn{1}{c}{UCF-RC} & \multicolumn{1}{c}{Charades-MLC}\\
         \midrule
           \addlinespace[0.01cm]
         None                     & 77.3  && 57.1   & 89.8   && 38.3  & 22.7    && 46.6   & 82.3 && 0.217 & 7.9  \\
           \addlinespace[0.01cm]
         \midrule
        \addlinespace[0.01cm]
         MoCo                     & 83.3  && 57.1   & 90.7   && 60.4   & 30.9   && 65.0   & 84.5   && 0.208   & 8.3  \\
         VideoMoCo                & 84.9  && 59.0   & 90.3   && 65.4   & 20.6   && 57.3   & 83.9   && 0.185   & 10.5  \\
         SeLaVi                   & 85.2  && 56.2   & 88.9   && 69.0   & 30.2   && 51.3   & 80.9   && 0.162   & 8.4  \\
         Pretext-Contrast         & 87.7  && 56.9   & 90.5   & &64.6   & 27.5   && 66.1   & 86.1   && 0.164   & 8.9  \\
         RSPNet                   & 88.7  && 59.0   & 91.1   && 74.7   & 32.2   && 65.4   & 83.6   && 0.145   & 9.0   \\
         AVID-CMA                 & 88.8  && 52.0   & 90.4   && 68.2   & 33.4   && 68.0   & 87.3   && 0.148   & 8.2   \\
         CtP                      & 90.1  && 59.6   & 92.0   && 61.0   & 32.9   && 79.1   & 88.8   && 0.178   & 9.6   \\
         TCLR                     & 90.8  && 59.8   & 91.6   && 72.6   & 26.3   && 60.7   & 84.7   && 0.142   & 12.2   \\
         GDT                      & 91.3  && 58.0   & 90.5   && 78.4   & 45.6   && 66.0   & 83.4   && 0.123   & 8.5   \\
        \addlinespace[0.01cm]
         \midrule
        \addlinespace[0.01cm]
         Supervised               & 93.9  && 60.8   & 92.1   && 86.6   & 51.3   && 79.0   & 87.1   && 0.132   & 23.5   \\
        \addlinespace[0.01cm]
         \bottomrule
    \end{tabular}
    }

    \label{proposed-benchmarks}
\end{table}

\section{Observations, Limitations and Recommendations}
\begin{figure}[t!]
 \captionsetup{font=small,skip=1mm}
     \centering
     \includegraphics[width=\linewidth]{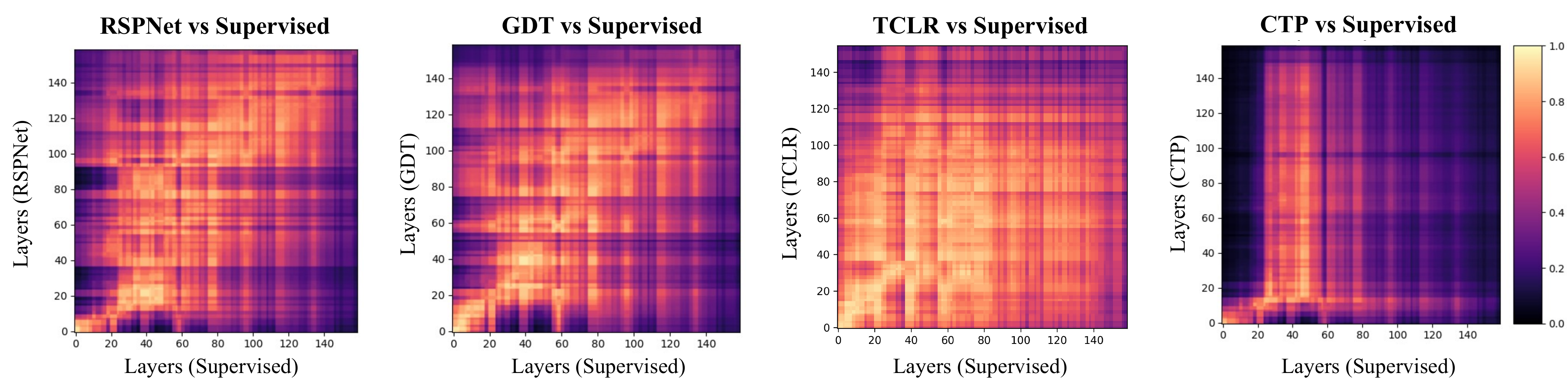}
     \caption{\textbf{Representation similarity} between features of top self-supervised methods and supervised pre-training on Kinetics-400 validation set (using centered kernel alignment~\cite{cka}). Contrastive methods have a high correlation with supervised pretraining, while CtP's features are far away. Thus, showing potential for both imitating supervised learning as well as learning features distinct to it.    
     }
     \label{features}
 \end{figure}

\textbf{Observations.}
We hope that our study and resulting benchmark provides a helpful insight for future research to design novel self-supervised methods for generalizable video representation learning. From the benchmark results in~\cref{proposed-benchmarks}, we observe that: 
\begin{enumerate}[label=(\roman*)]
\item There is no clear winner as different methods stand out in different downstream settings.

\item Supervised pre-training is dominant across all sensitivity factors, especially when the number of available downstream samples are limited and when there is a change in both the downstream domain and the downstream task.

\item Self-supervised contrastive methods that explicitly encourage features to be distinct across the temporal dimension transfer well. This is visible from the consistent performance of GDT, TCLR and RSPNet across different sensitivity factors. 

\item Learning certain temporal invariances may prevent generalizability to temporal or fine-grained benchmarks.
This is evident from GDT's performance on SS-v2 and UB-S1. These benchmarks require distinction between actions such as \textit{moving something left} vs. \textit{moving something right} in SS-v2 and \textit{giant circle forwards} vs. \textit{giant circle backwards} in UB-S1. The invariance to temporal reversal learned by GDT impacts its ability to recognize such actions. Similarly, MoCo outperforming VideoMoCo on the FX-S1 and UB-S1 Gym-99 subsets suggests that invariance to frame dropout in VideMoCo can harm the performance on highly similar actions.

\item Pretext-tasks specific to videos can be effective to learn more fine-grained features. CtP generalizes well both to different domains where the background is less indicative of the action and to more semantically similar actions. The pretext task is to track and estimate the position and size of image patches moving in a sequence of video frames. Such a formulation requires the network to learn to follow moving targets and ignore the static background information. CtP's generalization success demonstrates that contrastive learning is not the only way forward for self-supervised video representation learning. 

\item \cref{features} shows the feature similarity on Kinetics using centered kernel alignment ~\cite{cka} between supervised pre-training and the best self-supervised methods~\ie GDT, RSPNet, TCLR, CtP.  
This figure illustrates that contrastive methods seem to imitate supervised pre-training as the correlation  between supervised pre-training and the three contrastive methods (RSPNet, GDT and TCLR) is high. This explains the good performance of these methods on UCF-101 with 1000 examples. By contrast, CtP's features are far away from supervised pre-training. This is interesting because CtP generalizes well to new domains and actions, it shows that good generalization capability can be obtained without imitating supervised pre-training. 
\end{enumerate}
\medskip
\noindent\textbf{Limitations.} While our study has highlighted the benchmark sensitivity of video self-supervised learning across four factors, there are many more factors that we do not consider in this work. %There are certain limitations to our study: 
Due to computational limits, we keep the source dataset fixed as Kinetics-400 and use publicly available pre-trained models. This means there is variability in the exact pre-training setup such as the spatial data augmentations that are used by each model. We hope that future works will explore impact of such pretraining factors as well as the impact of pre-training on other large-scale datasets such as Ego4D~\cite{Ego4D2021} for the generalization of video self-supervised models. Another limitation of our study is that we only consider a fixed R(2+1)D-18 backbone, which is currently one of the most commonly used in video self-supervised learning. This allows our comparison between methods to be fair, however, it does limit the ability of methods to perform well on datasets such as EPIC-Kitchens-100. Another factor that could be explored further is the task. We have considered a selection of various video understanding tasks centered around human actions. However, there are many more video understanding tasks that could be explored such as human centric tasks like action anticipation~\cite{EPIC-100-arxiv} and temporal action detection\cite{EPIC-100-arxiv}, as well as non-human centric tasks like animal behavior analysis \cite{animal1,animal2,animal3}, multi-object tracking~\cite{pedersen20203d} and visual grounding~\cite{animal3}.

\medskip
\noindent\textbf{Recommendations.} Based on the results and our observations, we have several recommendations for future works in video self-supervised learning. 
(i) Our study has highlighted the need for more focus on generalizability of self-supervised learning methods, particularly along the domain and dataset size factors.
(ii) Distinguishing across the temporal dimension is effective and is a useful direction to pursue further for generalizability.
(iii) Pretext-tasks like the one used in CtP are good for the generalizability to domain and action, thus designing new video specific pretext tasks is a promising direction. This could also be combined with contrastive learning tasks to gain the benefits of both types of learning.

% % default ECCV template stuff
% \import{sections/}{eccv_default.tex}
%\clearpage
\medskip
\noindent\textbf{Acknowledgements.}
This work is part of the research programme Perspectief EDL with project number P16-25 project 3, which is financed by the Dutch Research Council (NWO) domain Applied and Engineering/ Sciences (TTW).
% ---- Bibliography ----
%
% BibTeX users should specify bibliography style 'splncs04'.
% References will then be sorted and formatted in the correct style.
%
\bibliographystyle{splncs04}
\bibliography{egbib}

\begin{thebibliography}{10}
\providecommand{\url}[1]{\texttt{#1}}
\providecommand{\urlprefix}{URL }
\providecommand{\doi}[1]{https://doi.org/#1}

\bibitem{mmvssl3-Afouras20b}
Afouras, T., Owens, A., Chung, J.S., Zisserman, A.: Self-supervised learning of
  audio-visual objects from video. In: European Conference on Computer Vision
  (ECCV) (2020)

\bibitem{jigaw1-ahsan2019video}
Ahsan, U., Madhok, R., Essa, I.: Video jigsaw: Unsupervised learning of
  spatiotemporal context for video action recognition. In: Proceedings of the
  IEEE Winter Conference on Applications of Computer Vision (WACV). pp.
  179--189. IEEE (2019)

\bibitem{alwassel2020self}
Alwassel, H., Mahajan, D., Korbar, B., Torresani, L., Ghanem, B., Tran, D.:
  Self-supervised learning by cross-modal audio-video clustering. In: Advances
  in Neural Information Processing Systems (NeurIPS). vol.~33, pp. 9758--9770
  (2020)

\bibitem{selavi-asano2020labelling}
Asano, Y.M., Patrick, M., Rupprecht, C., Vedaldi, A.: Labelling unlabelled
  videos from scratch with multi-modal self-supervision. In: Advances in Neural
  Information Processing Systems (NeurIPS) (2020)

\bibitem{asano2019critical}
Asano, Y.M., Rupprecht, C., Vedaldi, A.: A critical analysis of
  self-supervision, or what we can learn from a single image. In: International
  Conference on Learning Representations (ICLR) (2020)

\bibitem{taco-bai2020can}
Bai, Y., Fan, H., Misra, I., Venkatesh, G., Lu, Y., Zhou, Y., Yu, Q., Chandra,
  V., Yuille, A.: Can temporal information help with contrastive
  self-supervised learning? arXiv preprint arXiv:2011.13046  (2020)

\bibitem{relative-speed1-benaim2020speednet}
Benaim, S., Ephrat, A., Lang, O., Mosseri, I., Freeman, W.T., Rubinstein, M.,
  Irani, M., Dekel, T.: Speednet: Learning the speediness in videos. In:
  Proceedings of the IEEE/CVF Conference on Computer Vision and Pattern
  Recognition (CVPR). pp. 9922--9931 (2020)

\bibitem{multimodal-clustering1-chen2021multimodal}
Chen, B., Rouditchenko, A., Duarte, K., Kuehne, H., Thomas, S., Boggust, A.,
  Panda, R., Kingsbury, B., Feris, R., Harwath, D., et~al.: Multimodal
  clustering networks for self-supervised learning from unlabeled videos. In:
  Proceedings of the IEEE/CVF International Conference on Computer Vision
  (CVPR). pp. 8012--8021 (2021)

\bibitem{simclr-pmlr-v119-chen20j}
Chen, T., Kornblith, S., Norouzi, M., Hinton, G.: A simple framework for
  contrastive learning of visual representations. In: Proceedings of the
  International Conference on Machine Learning (PMLR) (2020)

\bibitem{moco_v2}
Chen, X., Fan, H., Girshick, R., He, K.: Improved baselines with momentum
  contrastive learning. arXiv preprint arXiv:2003.04297  (2020)

\bibitem{playback1-cho2020self}
Cho, H., Kim, T., Chang, H.J., Hwang, W.: Self-supervised spatio-temporal
  representation learning using variable playback speed prediction. IEEE Access
   \textbf{9},  79562--79571 (2021)

\bibitem{when_contrast_work}
Cole, E., Yang, X., Wilber, K., Mac~Aodha, O., Belongie, S.: When does
  contrastive visual representation learning work? In: Proceedings of the
  IEEE/CVF Conference on Computer Vision and Pattern Recognition (CVPR) (2022)

\bibitem{EPIC-100-arxiv}
Damen, D., Doughty, H., Farinella, G.M., , Furnari, A., Ma, J., Kazakos, E.,
  Moltisanti, D., Munro, J., Perrett, T., Price, W., Wray, M.: Rescaling
  egocentric vision: Collection, pipeline and challenges for
  {EPIC-KITCHENS-100}. International Journal of Computer Vision (IJCV)  (2021)

\bibitem{dave2021tclr}
Dave, I., Gupta, R., Rizve, M.N., Shah, M.: Tclr: Temporal contrastive learning
  for video representation. In Computer Vision and Image Understanding (CVIU)
  p. 103406 (2022)

\bibitem{diba2021vi2clr}
Diba, A., Sharma, V., Safdari, R., Lotfi, D., Sarfraz, S., Stiefelhagen, R.,
  Van~Gool, L.: Vi2clr: Video and image for visual contrastive learning of
  representation. In: Proceedings of the IEEE/CVF International Conference on
  Computer Vision (ICCV). pp. 1502--1512 (2021)

\bibitem{edinburgh-ericsson2021well}
Ericsson, L., Gouk, H., Hospedales, T.M.: How well do self-supervised models
  transfer? In: Proceedings of the IEEE/CVF Conference on Computer Vision and
  Pattern Recognition (CVPR). pp. 5414--5423 (2021)

\bibitem{ericsson2021self}
Ericsson, L., Gouk, H., Hospedales, T.M.: Why do self-supervised models
  transfer? investigating the impact of invariance on downstream tasks. arXiv
  preprint arXiv:2111.11398  (2021)

\bibitem{animal1}
Eyjolfsdottir, E., Branson, S., Burgos-Artizzu, X.P., Hoopfer, E.D., Schor, J.,
  Anderson, D.J., Perona, P.: Detecting social actions of fruit flies. In:
  European Conference on Computer Vision. pp. 772--787 (2014)

\bibitem{Feichtenhofer2019SlowFastNF}
Feichtenhofer, C., Fan, H., Malik, J., He, K.: Slowfast networks for video
  recognition. In: Proceedings of the IEEE/CVF International Conference on
  Computer Vision (ICCV). pp. 6201--6210 (2019)

\bibitem{large-scale-feichtenhofer2021large}
Feichtenhofer, C., Fan, H., Xiong, B., Girshick, R., He, K.: A large-scale
  study on unsupervised spatiotemporal representation learning. In: Proceedings
  of the IEEE/CVF Conference on Computer Vision and Pattern Recognition (CVPR).
  pp. 3299--3309 (2021)

\bibitem{shuffle1-fernando2017self}
Fernando, B., Bilen, H., Gavves, E., Gould, S.: Self-supervised video
  representation learning with odd-one-out networks. In: Proceedings of the
  IEEE Conference on Computer Vision and Pattern Recognition (CVPR). pp.
  3636--3645 (2017)

\bibitem{gavrilyuk2021motion}
Gavrilyuk, K., Jain, M., Karmanov, I., Snoek, C.G.M.: Motion-augmented
  self-training for video recognition at smaller scale. In: Proceedings of the
  IEEE/CVF International Conference on Computer Vision (ICCV). pp. 10429--10438
  (2021)

\bibitem{arrow_of_time}
Ghodrati, A., Gavves, E., Snoek, C.G.M.: Video time: Properties, encoders and
  evaluation. In: British Machine Vision Conference (BMVC) (2018)

\bibitem{goyal2019scaling}
Goyal, P., Mahajan, D., Gupta, A., Misra, I.: Scaling and benchmarking
  self-supervised visual representation learning. In: Proceedings of the
  IEEE/CVF International Conference on Computer Vision (ICCV). pp. 6391--6400
  (2019)

\bibitem{SS-v2-arxiv}
Goyal, R., Ebrahimi~Kahou, S., Michalski, V., Materzynska, J., Westphal, S.,
  Kim, H., Haenel, V., Fruend, I., Yianilos, P., Mueller-Freitag, M., et~al.:
  The ``something something" video database for learning and evaluating visual
  common sense. In: Proceedings of the IEEE International Conference on
  Computer Vision (ICCV). pp. 5842--5850 (2017)

\bibitem{Ego4D2021}
Grauman, K., et~al.: Ego4d: Around the {W}orld in 3,000 {H}ours of {E}gocentric
  {V}ideo. In: Proceedings of the IEEE/CVF Conference on Computer Vision and
  Pattern Recognition (CVPR) (2022)

\bibitem{AVA-Gu_2018_CVPR}
Gu, C., Sun, C., Ross, D.A., Vondrick, C., Pantofaru, C., Li, Y.,
  Vijayanarasimhan, S., Toderici, G., Ricco, S., Sukthankar, R., Schmid, C.,
  Malik, J.: Ava: A video dataset of spatio-temporally localized atomic visual
  actions. In: Proceedings of the IEEE Conference on Computer Vision and
  Pattern Recognition (CVPR) (2018)

\bibitem{han2019video}
Han, T., Xie, W., Zisserman, A.: Video representation learning by dense
  predictive coding. In: Proceedings of the IEEE/CVF International Conference
  on Computer Vision Workshops (2019)

\bibitem{coclr}
Han, T., Xie, W., Zisserman, A.: Self-supervised co-training for video
  representation learning. In: Advances in Neural Information Processing
  Systems (NeurIPS) (2020)

\bibitem{multimodal-clustering2-hu2019deep}
Hu, D., Nie, F., Li, X.: Deep multimodal clustering for unsupervised
  audiovisual learning. In: Proceedings of the IEEE/CVF Conference on Computer
  Vision and Pattern Recognition (CVPR). pp. 9248--9257 (2019)

\bibitem{huang2021ascnet}
Huang, D., Wu, W., Hu, W., Liu, X., He, D., Wu, Z., Wu, X., Tan, M., Ding, E.:
  Ascnet: Self-supervised video representation learning with appearance-speed
  consistency. In: Proceedings of the IEEE/CVF International Conference on
  Computer Vision (ICCV). pp. 8096--8105 (2021)

\bibitem{jigsaw2-huo2021selfsupervised}
Huo, Y., Ding, M., Lu, H., Lu, Z., Xiang, T., Wen, J.R., Huang, Z., Jiang, J.,
  Zhang, S., Tang, M., Huang, S., Luo, P.: Self-supervised video representation
  learning with constrained spatiotemporal jigsaw. In: Proceedings of the
  Thirtieth International Joint Conference on Artificial Intelligence (IJCAI)
  (2021)

\bibitem{trasnferability}
Islam, A., Chen, C.F.R., Panda, R., Karlinsky, L., Radke, R., Feris, R.: A
  broad study on the transferability of visual representations with contrastive
  learning. In: Proceedings of the IEEE/CVF International Conference on
  Computer Vision (ICCV). pp. 8845--8855 (2021)

\bibitem{relative-speed2-jenni2020video}
Jenni, S., Meishvili, G., Favaro, P.: Video representation learning by
  recognizing temporal transformations. In: European Conference on Computer
  Vision (ECCV). pp. 425--442 (2020)

\bibitem{rotate1-jing2019selfsupervised}
Jing, L., Yang, X., Liu, J., Tian, Y.: Self-supervised spatiotemporal feature
  learning via video rotation prediction. arXiv preprint arXiv:1811.11387
  (2018)

\bibitem{Kinetics-400-arxiv}
Kay, W., Carreira, J., Simonyan, K., Zhang, B., Hillier, C., Vijayanarasimhan,
  S., Viola, F., Green, T., Back, T., Natsev, P., Suleyman, M., Zisserman, A.:
  The kinetics human action video dataset. arXiv preprint arXiv:1705.06950
  (2017)

\bibitem{jigsaw3-kim2019self}
Kim, D., Cho, D., Kweon, I.S.: Self-supervised video representation learning
  with space-time cubic puzzles. In: Proceedings of the AAAI Conference on
  Artificial Intelligence. pp. 8545--8552 (2019)

\bibitem{kolesnikov2019revisiting}
Kolesnikov, A., Zhai, X., Beyer, L.: Revisiting self-supervised visual
  representation learning. In: Proceedings of the IEEE/CVF Conference on
  Computer Vision and Pattern Recognition (CVPR). pp. 1920--1929 (2019)

\bibitem{yowo}
K{\"{o}}p{\"{u}}kl{\"{u}}, O., Wei, X., Rigoll, G.: You only watch once: {A}
  unified {CNN} architecture for real-time spatiotemporal action localization.
  arXiv preprint arXiv:1911.06644  (2019)

\bibitem{korbar2018cooperative}
Korbar, B., Tran, D., Torresani, L.: Cooperative learning of audio and video
  models from self-supervised synchronization. In: Advances in Neural
  Information Processing Systems (NeurIPS). vol.~31 (2018)

\bibitem{contrasting_contrastive}
Kotar, K., Ilharco, G., Schmidt, L., Ehsani, K., Mottaghi, R.: Contrasting
  contrastive self-supervised representation learning pipelines. In:
  Proceedings of the IEEE/CVF International Conference on Computer Vision
  (ICCV). pp. 9949--9959 (2021)

\bibitem{HMDB-51-ICCV}
Kuehne, H., Jhuang, H., Garrote, E., Poggio, T., Serre, T.: {HMDB}: a large
  video database for human motion recognition. In: Proceedings of the
  International Conference on Computer Vision (ICCV) (2011)

\bibitem{diving}
Li, Y., Li, Y., Vasconcelos, N.: Resound: Towards action recognition without
  representation bias. In: Proceedings of the European Conference on Computer
  Vision (ECCV). pp. 513--528 (2018)

\bibitem{lin2021self}
Lin, Y., Guo, X., Lu, Y.: Self-supervised video representation learning with
  meta-contrastive network. In: Proceedings of the IEEE/CVF International
  Conference on Computer Vision (ICCV). pp. 8239--8249 (2021)

\bibitem{vcp}
Luo, D., Liu, C., Zhou, Y., Yang, D., Ma, C., Ye, Q., Wang, W.: Video cloze
  procedure for self-supervised spatio-temporal learning. In: Proceedings of
  the AAAI Conference on Artificial Intelligence. pp. 11701--11708 (2020)

\bibitem{ma2021active}
Ma, S., Zeng, Z., McDuff, D., Song, Y.: Active contrastive learning of
  audio-visual video representations. In: International Conference on Learning
  Representations (ICLR) (2021)

\bibitem{mettes2016spot}
Mettes, P., Gemert, J.C.v., Snoek, C.G.M.: Spot on: Action localization from
  pointly-supervised proposals. In: European conference on computer vision
  (ECCV). pp. 437--453. Springer (2016)

\bibitem{frame-order-misra2016shuffle}
Misra, I., Zitnick, C.L., Hebert, M.: Shuffle and learn: unsupervised learning
  using temporal order verification. In: European Conference on Computer Vision
  (ECCV). pp. 527--544 (2016)

\bibitem{avid-cma-morgado2021audio}
Morgado, P., Vasconcelos, N., Misra, I.: Audio-visual instance discrimination
  with cross-modal agreement. In: Proceedings of the IEEE/CVF Conference on
  Computer Vision and Pattern Recognition (CVPR) (2021)

\bibitem{newell2020useful}
Newell, A., Deng, J.: How useful is self-supervised pretraining for visual
  tasks? In: Proceedings of the IEEE/CVF Conference on Computer Vision and
  Pattern Recognition (CVPR) (2020)

\bibitem{animal2}
Ng, X.L., Ong, K.E., Zheng, Q., Ni, Y., Yeo, S.Y., Liu, J.: Animal kingdom: A
  large and diverse dataset for animal behavior understanding. In: Proceedings
  of the IEEE/CVF Conference on Computer Vision and Pattern Recognition (CVPR).
  pp. 19023--19034 (2022)

\bibitem{cka}
Nguyen, T., Raghu, M., Kornblith, S.: Do wide and deep networks learn the same
  things? uncovering how neural network representations vary with width and
  depth. In: International Conference on Learning Representations (ICLR) (2021)

\bibitem{videomoco-pan2021videomoco}
Pan, T., Song, Y., Yang, T., Jiang, W., Liu, W.: Videomoco: Contrastive video
  representation learning with temporally adversarial examples. In: Proceedings
  of the IEEE/CVF Conference on Computer Vision and Pattern Recognition (CVPR).
  pp. 11205--11214 (2021)

\bibitem{gdt-patrick2020multimodal}
Patrick, M., Asano, Y.M., Kuznetsova, P., Fong, R., Henriques, J.F., Zweig, G.,
  Vedaldi, A.: Multi-modal self-supervision from generalized data
  transformations. In: International Conference on Computer Vision (ICCV)
  (2021)

\bibitem{pedersen20203d}
Pedersen, M., Haurum, J.B., Bengtson, S.H., Moeslund, T.B.: 3d-zef: A 3d
  zebrafish tracking benchmark dataset. In: Proceedings of the IEEE/CVF
  Conference on Computer Vision and Pattern Recognition. pp. 2426--2436 (2020)

\bibitem{rspnet-chen2020RSPNet}
Peihao, C., Deng, H., Dongliang, H., Xiang, L., Runhao, Z., Shilei, W.,
  Mingkui, T., Chuang, G.: Rspnet: Relative speed perception for unsupervised
  video representation learning. In: The AAAI Conference on Artificial
  Intelligence (AAAI) (2021)

\bibitem{piergiovanni2020evolving}
Piergiovanni, A., Angelova, A., Ryoo, M.S.: Evolving losses for unsupervised
  video representation learning. In: Proceedings of the IEEE/CVF Conference on
  Computer Vision and Pattern Recognition (CVPR). pp. 133--142 (2020)

\bibitem{qian2021spatiotemporal}
Qian, R., Meng, T., Gong, B., Yang, M.H., Wang, H., Belongie, S., Cui, Y.:
  Spatiotemporal contrastive video representation learning. In: Proceedings of
  the IEEE/CVF Conference on Computer Vision and Pattern Recognition (CVPR).
  pp. 6964--6974 (2021)

\bibitem{recasens2021broaden}
Recasens, A., Luc, P., Alayrac, J.B., Wang, L., Strub, F., Tallec, C.,
  Malinowski, M., P{\u{a}}tr{\u{a}}ucean, V., Altch{\'e}, F., Valko, M.,
  et~al.: Broaden your views for self-supervised video learning. In:
  Proceedings of the IEEE/CVF International Conference on Computer Vision
  (ICCV). pp. 1255--1265 (2021)

\bibitem{sariyildiz2021concept}
Sariyildiz, M.B., Kalantidis, Y., Larlus, D., Alahari, K.: Concept
  generalization in visual representation learning. In: Proceedings of the
  IEEE/CVF International Conference on Computer Vision (ICCV). pp. 9629--9639
  (2021)

\bibitem{schiappa2022self}
Schiappa, M.C., Rawat, Y.S., Shah, M.: Self-supervised learning for videos: A
  survey. arXiv preprint arXiv:2207.00419  (2022)

\bibitem{NTU-60-arxiv}
Shahroudy, A., Liu, J., Ng, T.T., Wang, G.: Ntu rgb+ d: A large scale dataset
  for 3d human activity analysis. In: Proceedings of the IEEE Conference on
  Computer Vision and Pattern Recognition (CVPR). pp. 1010--1019 (2016)

\bibitem{Gym-99-arxiv}
Shao, D., Zhao, Y., Dai, B., Lin, D.: Finegym: A hierarchical video dataset for
  fine-grained action understanding. In: Proceedings of the IEEE Conference on
  Computer Vision and Pattern Recognition (CVPR) (2020)

\bibitem{charades-sigurdsson:hal-01418216}
Sigurdsson, G.A., Varol, G., Wang, X., Farhadi, A., Laptev, I., Gupta, A.:
  {Hollywood in Homes: Crowdsourcing Data Collection for Activity
  Understanding}. In: European Conference on Computer Vision (ECCV). pp. 510 --
  526 (2016)

\bibitem{UCF-101-arxiv}
Soomro, K., Zamir, A.R., Shah, M.: Ucf101: A dataset of 101 human actions
  classes from videos in the wild. arXiv preprint arXiv:1212.0402  (2012)

\bibitem{sun2021composable}
Sun, C., Nagrani, A., Tian, Y., Schmid, C.: Composable augmentation encoding
  for video representation learning. In: Proceedings of the IEEE/CVF
  International Conference on Computer Vision (ICCV). pp. 8834--8844 (2021)

\bibitem{animal3}
Sun, J.J., Karigo, T., Chakraborty, D., Mohanty, S., Wild, B., Sun, Q., Chen,
  C., Anderson, D., Perona, P., Yue, Y., Kennedy, A.: The multi-agent behavior
  dataset: Mouse dyadic social interactions. In: Vanschoren, J., Yeung, S.
  (eds.) Proceedings of the Neural Information Processing Systems Track on
  Datasets and Benchmarks (2021)

\bibitem{shuffle2-suzuki2018learning}
Suzuki, T., Itazuri, T., Hara, K., Kataoka, H.: Learning spatiotemporal 3d
  convolution with video order self-supervision. In: Proceedings of the
  European Conference on Computer Vision (ECCV) Workshops. pp.~0--0 (2018)

\bibitem{tao2020self}
Tao, L., Wang, X., Yamasaki, T.: Self-supervised video representation learning
  using inter-intra contrastive framework. In: Proceedings of the 28th ACM
  International Conference on Multimedia (ACM MM. pp. 2193--2201 (2020)

\bibitem{pretext-contrast-DBLP:journals/corr/abs-2010-15464}
Tao, L., Wang, X., Yamasaki, T.: Pretext-contrastive learning: Toward good
  practices in self-supervised video representation leaning. arXiv preprint
  arXiv:2010.15464  (2021)

\bibitem{fmthoker_acmmm}
Thoker, F.M., Doughty, H., Snoek, C.: Skeleton-contrastive 3d action
  representation learning. In: in Proceedings of the 29th ACM International
  Conference on Multimedia,(ACM MM ) (2021)

\bibitem{tran2018closer}
Tran, D., Wang, H., Torresani, L., Ray, J., LeCun, Y., Paluri, M.: A closer
  look at spatiotemporal convolutions for action recognition. In: Proceedings
  of the IEEE Conference on Computer Vision and Pattern Recognition (CVPR). pp.
  6450--6459 (2018)

\bibitem{van2021benchmarking}
Van~Horn, G., Cole, E., Beery, S., Wilber, K., Belongie, S., Mac~Aodha, O.:
  Benchmarking representation learning for natural world image collections. In:
  Proceedings of the IEEE/CVF Conference on Computer Vision and Pattern
  Recognition (CVPR). pp. 12884--12893 (2021)

\bibitem{image-eval5-wallace2020extending}
Wallace, B., Hariharan, B.: Extending and analyzing self-supervised learning
  across domains. In: European Conference on Computer Vision (ECCV). pp.
  717--734. Springer (2020)

\bibitem{ctp-wang2021unsupervised}
Wang, G., Zhou, Y., Luo, C., Xie, W., Zeng, W., Xiong, Z.: Unsupervised visual
  representation learning by tracking patches in video. In: Proceedings of the
  IEEE Conference on Computer Vision and Pattern Recognition (CVPR) (2021)

\bibitem{wang2019self}
Wang, J., Jiao, J., Bao, L., He, S., Liu, Y., Liu, W.: Self-supervised
  spatio-temporal representation learning for videos by predicting motion and
  appearance statistics. In: Proceedings of the IEEE/CVF Conference on Computer
  Vision and Pattern Recognition (CVPR). pp. 4006--4015 (2019)

\bibitem{playback3-wang2020self}
Wang, J., Jiao, J., Liu, Y.H.: Self-supervised video representation learning by
  pace prediction. In: European Conference on Computer Vision (ECCV). pp.
  504--521 (2020)

\bibitem{wang2021removing}
Wang, J., Gao, Y., Li, K., Lin, Y., Ma, A.J., Cheng, H., Peng, P., Ji, R., Sun,
  X.: Removing the background by adding the background: Towards background
  robust self-supervised video representation learning. In: Proceedings of the
  IEEE/CVF Conference on Computer Vision and Pattern Recognition (CVPR) (2021)

\bibitem{aot2-wei2018learning}
Wei, D., Lim, J.J., Zisserman, A., Freeman, W.T.: Learning and using the arrow
  of time. In: Proceedings of the IEEE Conference on Computer Vision and
  Pattern Recognition (CVPR). pp. 8052--8060 (2018)

\bibitem{xiao2021modist}
Xiao, F., Tighe, J., Modolo, D.: Modist: Motion distillation for
  self-supervised video representation learning. arXiv preprint
  arXiv:2106.09703  (2021)

\bibitem{clip-order-xu2019self}
Xu, D., Xiao, J., Zhao, Z., Shao, J., Xie, D., Zhuang, Y.: Self-supervised
  spatiotemporal learning via video clip order prediction. In: Proceedings of
  the IEEE/CVF Conference on Computer Vision and Pattern Recognition (CVPR).
  pp. 10334--10343 (2019)

\bibitem{yang2020video}
Yang, C., Xu, Y., Dai, B., Zhou, B.: Video representation learning with visual
  tempo consistency. arXiv preprint arXiv:2006.15489  (2020)

\bibitem{yang2020transfer}
Yang, X., He, X., Liang, Y., Yang, Y., Zhang, S., Xie, P.: Transfer learning or
  self-supervised learning? a tale of two pretraining paradigms. arXiv preprint
  arXiv:2007.04234  (2020)

\bibitem{yao2021seco}
Yao, T., Zhang, Y., Qiu, Z., Pan, Y., Mei, T.: Seco: Exploring sequence
  supervision for unsupervised representation learning. In: AAAI. vol.~2, p.~7
  (2021)

\bibitem{playback2-yao2020video}
Yao, Y., Liu, C., Luo, D., Zhou, Y., Ye, Q.: Video playback rate perception for
  self-supervised spatio-temporal representation learning. In: Proceedings of
  the IEEE/CVF Conference on Computer Vision and Pattern Recognition (CVPR).
  pp. 6548--6557 (2020)

\bibitem{zhai2019large}
Zhai, X., Puigcerver, J., Kolesnikov, A., Ruyssen, P., Riquelme, C., Lucic, M.,
  Djolonga, J., Pinto, A.S., Neumann, M., Dosovitskiy, A., et~al.: A
  large-scale study of representation learning with the visual task adaptation
  benchmark. arXiv preprint arXiv:1910.04867  (2019)

\bibitem{rep_counting}
Zhang, H., Xu, X., Han, G., He, S.: Context-aware and scale-insensitive
  temporal repetition counting. In: Proceedings of the IEEE/CVF Conference on
  Computer Vision and Pattern Recognition (CVPR) (2020)

\bibitem{pretext-contrast-2-zhang2021contrastive}
Zhang, Y., Po, L.M., Xu, X., Liu, M., Wang, Y., Ou, W., Zhao, Y., Yu, W.Y.:
  Contrastive spatio-temporal pretext learning for self-supervised video
  representation. In: Proceedings of the AAAI Conference on Artificial
  Intelligenc (2022)

\end{thebibliography}

\clearpage

\appendix

%\title{
%}{\big{How Severe is Benchmark-Sensitivity in Video Self-Supervised Learning?}}
%\noindent\textbf{\large{How Severe is Benchmark-Sensitivity in Video Self-Supervised Learning?}}\\

\noindent\textbf{\Large{Appendix}}

\bigskip

 In \cref{sec:evaluated-models}, we provide details of the video self-supervised models we use in our evaluation study. \cref{sec:expt-details} provides details on the experimental setup for each of our downstream sensitivity factors. We also show correlation plots between current benchmarks and the experimental results for each sensitivity factor in \cref{correlattion_plots}. Feature similarities between supervised pre-training and each self-supervised pre-training method are shown in \cref{similarity_features}. In \cref{sec:video-datasets}, we describe domain difference between the downstream video datasets we use and the attributes we use to characterize this difference. We show the standard deviations of the experiments on the SEVERE benchmark \cref{sec:std_dev} and also compare the SEVERE benchmark to results on HMDB51 action recognition in \cref{hmdb}. Finally, we report results of some  additional experiments in \cref{sec:lin_eval_samples} and \cref{sec:epic_nouns} that we did not have room for in the main paper.% due to the lack of space.

\section{Details of the Evaluated Self-Supervised Models}
\label{sec:evaluated-models}
We use a variety of different self-supervised methods in our paper, here we describe each method:

\noindent\textbf{MoCo~\cite{moco_v2}} is a contrastive learning method proposed for representation learning in images. Positives are created by performing different spatial augmentations on a video. Negatives are other videos. To obtain negatives beyond the current batch, MoCo proposes a momentum encoder which maintains a queue of momentum-updated data samples from previous batches.

\noindent\textbf{SeLaVi~\cite{selavi-asano2020labelling}} views the audio and visual modalities as different augmentations of a video and learns with a cross-modal clustering pretext task.

\noindent\textbf{VideoMoCo~\cite{videomoco-pan2021videomoco}} extends MoCo to the temporal domain. It does this with an adversarial dropout augmentation which removes the frames the model considers most important. With the contrastive learning loss, the model learns invariance to this adversarial frame dropout alongside the spatial augmentations used in MoCo.

\noindent\textbf{Pretext-Contrast~\cite{pretext-contrast-DBLP:journals/corr/abs-2010-15464}} combines the pretext task approach with contrastive learning. As its pretext task it uses video cloze procedure~\cite{vcp} where the goal is to predict which augmentations have been applied to a video clip. For the contrastive learning objective different temporal shifts, \ie distinct clips from the same video, are considered.

\noindent\textbf{RSPNet~\cite{rspnet-chen2020RSPNet}} also combines pretext and contrastive tasks, with a focus on video speed. The pretext task is to predict the relative difference in speed between two versions of the same video, while the contrastive task creates extra positives and negatives by augmenting videos with different speeds along with the spatial augmentations.

\noindent\textbf{AVID-CMA~\cite{avid-cma-morgado2021audio}} is a multi-modal contrastive learning method which uses audio in addition to the visual modality. It first uses cross-modal contrastive learning where the one modality is used as the positives and the other as the negatives. Then it uses within modality contrastive learning where additional positives which have high audio and visual similarity are sampled.

\noindent\textbf{CtP~\cite{ctp-wang2021unsupervised}} performs self-supervised learning through a ``catch the patch'' pretext task. The goal in this task is to predict the trajectory of an image patch which is resized and moved through a sequence of video frames.

\noindent\textbf{TCLR~\cite{dave2021tclr}} is a contrastive method which encourages features to be distinct across the temporal dimension. It does this by using clips from the same video as negatives. Therefore, instead of encouraging invariance to temporal shift as other methods to, it encourages the model to be able to distinguish between different shifts. It also uses an extensive set of spatial augmentations.

\noindent\textbf{GDT~\cite{gdt-patrick2020multimodal}} is a multi-modal contrastive method which composes a series of different augmentations and encourages model to learn invariance to some and learns to distinguish between others. We use the best performing version of GDT which encourages invariance to spatial augmentations, the audio and visual modalities and temporal %shift and 
reversal, while encouraging the model to distinguish between different temporal shifts.

While all models are pre-trained on Kinetics-400 and use an R(2+1)D-18 backbone with 112x112 spatial input size, there are some smaller differences in how the models are trained. Due to the computational cost of training these models we download publicly available models or obtain them from the authors, therefore we cannot control for these smaller differences in the pre-training set up. These differences include number of pre-training epochs, batch size, number of video frames used and spatial and temporal augmentations. We list these differences in Table~\ref{tab:method_diffs}.

\begin{table}[]
\captionsetup{font=footnotesize,skip=1mm}
\centering
\caption{\textbf{Pre-training differences of our evaluated self-supervised methods.} While all models are pre-trained with the same backbone and dataset, there are differences in how many epoches they were trained for, the batch size and number of frames they use and the spatial and temporal augmentations they are encouraged to be invariant to.}
\resizebox{\textwidth}{!}{%
\begin{tabular}{lllllccccccccccc}
\toprule
\multicolumn{1}{l}{\multirow{3}{*}{\textbf{Method}}} &
  &  & & && 
  \multicolumn{6}{c}{\textbf{Spatial Augmentations}} &
  \multicolumn{1}{c}{} &
  \multicolumn{3}{c}{\textbf{Temporal Augmentations}} \\
 \cmidrule{7-12} \cmidrule{14-16}
\multicolumn{1}{c}{} &  Extra & Epochs & Batch & Num & 
  \multicolumn{1}{c}{} &
  \multicolumn{1}{c}{Random} &
  \multicolumn{1}{c}{Horiz.} &
  \multicolumn{1}{c}{Grayscale} &
  \multicolumn{1}{c}{Color} &
  \multicolumn{1}{c}{Gaussian} &
  \multicolumn{1}{c}{Scaling} &
  \multicolumn{1}{c}{} &
  \multicolumn{1}{c}{Shift} &
  \multicolumn{1}{c}{Reversal} &
  \multicolumn{1}{c}{Speed} \\
  & Modality & & Size & Frames  & & Crop & Flip & & Jitter & Blur & \\
\midrule
MoCo & & 200 & 128 & 16 &  & \ding{51} & \ding{51} & \ding{51} & \ding{51} & & & &\ding{51}& & \\
SeLaVi &
  Audio & 200 & 1024 & 30 &  &
  \ding{51} &
  \ding{51} &
   &
   &
   &
   &
   &
   &
   &
    \\
  VideoMoCo & &
  200 & 128 & 32 &  & 
  \ding{51} &
  \ding{51} &
  \ding{51} &
  \ding{51} &
  &
  &
   &
  &
  & \\
  Pretext-Contrast &
   & 200 & 16 & 16 &  &
  \ding{51} &
  \ding{51} &
  \ding{51} &
  \ding{51} &
  \ding{51} &
   &
   &
  \ding{51} &
   &
    \\
RSPNet &
   & 200 & 64 & 16 & &
  \ding{51} &
   &
   &
  \ding{51} &
  \ding{51} &
   &
   &
  \ding{51} &
   &
  \ding{51} \\
  AVID-CMA &
  Audio & 400 & 256 & 16 &  &
  \ding{51} &
  \ding{51} &
   &
  \ding{51} &
   &
  \ding{51} &
   &
   &
 \\
 CtP &
   & 90 & 32 & 16  
 \\
TCLR &
   & 100 & 40 & 16&  &
  \ding{51} &
  \ding{51} &
  \ding{51} &
  \ding{51} &
   &
  \ding{51} &
   &
   &
   & \\
GDT &
  Audio & 100 & 512 & 30 &  &
  \ding{51} &
  \ding{51} &
   &\ding{51}
   &
   &
   &
   &
   &
  \ding{51} &
   \\
   \midrule
   Supervised & & 45 & 32 & 16 &  &
   \ding{51} & \ding{51} 
   &
   & 
   &
   &
   &
   &
   \ding{51}\\
\bottomrule
\end{tabular}%
}
\label{tab:method_diffs}
\end{table}

\section{Downstream Experimental Details}
\label{sec:expt-details}

\subsection{Downstream Domain}
\label{app:domain-shift-expt}
In \cref{sec:factor_1} we investigate to what extent self-supervised methods learn features applicable to action recognition in any domain. Here we explain the datasets, splits and training details we use to do this.

\medskip
\noindent\textbf{Datasets} We report our experiments on the following datasets:\\
\textit{UCF-101} \cite{UCF-101-arxiv} is currently one of the most widely used datasets for evaluating video self-supervised learning models. It consists of YouTube videos from a set of 101 coarse-grained classes with a high overlap with actions in Kinetics-400. We use the first standard split proposed in the original paper \cite{UCF-101-arxiv} containing 9,537 training and  3,783 testing samples for the 101 action classes.\\
\textit{NTU-60}: \cite{NTU-60-arxiv} consists of daily human actions captured in a controlled lab setting with a fixed number actors. Although it has some overlap with Kinetics-400 actions, it is quite different visually due to the setting. We use the cross-subject protocol proposed in \cite{NTU-60-arxiv} to split the data into 40,320 training and 16,560 testing samples for 60 action classes.\\
\textit{Gym-99}. We use FineGym version $v1.0$ \cite{Gym-99-arxiv} which is a dataset of fine-grained actions constructed from  recorded gymnastic competitions. We use the Gym 99 subset which contains 99 action classes with 20,484 and 8,521 samples in the train and test sets respectively.\\
\textit{SS-v2}: \cite{SS-v2-arxiv} is a crowdsourced collection of first-person videos aimed to
instill
common-sense understanding. It differs significantly with respect to Kinetics-400 in terms of visual appearance and point-of-view. We use the original dataset splits from \cite{SS-v2-arxiv} containing 168,913  training and 24,777 testing samples for 174 action classes.\\
\textit{EPIC-Kitchens-100}: \cite{EPIC-100-arxiv} is a large-scale egocentric dataset consisting of daily actions performed in a kitchen. It has annotations for verbs (97) and nouns (300) and the action is defined a tuple of these. Like SS-v2, EK-100 also differs significantly from Kinetics-400 in terms of visual appearance and point-of-view. We use standard splits from \cite{EPIC-100-arxiv} containing 67,217 samples in training set and 9,668 in the validation set. In the main paper we only aim to recognize the 97 verb classes, we provide results for the noun and action recognition tasks in \cref{sec:epic_nouns}.

\noindent\textbf{Training Details}
In the initial hyper-parameter search, we perform a grid search over various finetuning settings with learning rates between 0.1 - 0.00001, varying total training epochs, data augmentations, and schedulers. We choose the optimal hyper-parameters based on the performances of the pretraining models on the validation sets of each dataset for each downstream task.

During training, we sample a random clip from each video of 32 frames with standard augmentations \ie a random multi-scale crop of size 112x112, random horizontal flipping and color jittering. We train with the Adam optimizer. The learning rates, scheduling and total number of epochs vary across datasets and are shown in \cref{tab:downstream_domain_training}.  However, each model is trained with the same hyper-parameters for the corresponding dataset.  For inference, we use 10 linearly spaced clips of 32 frames each. For each frame we take a center crop which is resized to 112x112 pixels. To calculate the action class prediction of a video, we take the mean of the predictions from each clip and report top-1 accuracy.

\begin{table}[]
\captionsetup{font=footnotesize,skip=1mm}
\centering
\caption{\textbf{Training details} of finetuning and linear evaluation on various downstream datasets. Learning rate is scheduled using  a multip-step scheduler with $\gamma = 0.1$ at corresponding steps for each dataset. We  train all the models with same hyperparameters for the corresponding dataset.}
\resizebox{\textwidth}{!}{%
\begin{tabular}{lccccccccc}
\toprule
\multicolumn{1}{l}{\multirow{3}{*}{\textbf{Dataset}}} &
  \multicolumn{4}{c}{\textbf{Finetuning}} & 
  \multicolumn{1}{c}{} &
  \multicolumn{4}{c}{\textbf{Linear Evaluation}} \\
 \cmidrule{2-5} \cmidrule{7-10}
  %& Batch Size & Lr & Epochs & Lr steps \\
 & Batch Size & Learning rate & Epochs & Steps & & Batch Size & Learning rate & Epochs &  Steps \\
\midrule
%\multicolumn{5}{c}{Finetuning}\\
%\hdashline 
UCF-101 &  32&  0.0001  & 160 & [60,100,140] & & 64&  0.01 & 100 & [40,80] \\
NTU-60 &   32&  0.0001 & 180 &  [90, 140, 160]  & &  64&  0.01 & 120 & [40,80,100] \\
Gym-99 &   32&  0.0001 & 160 & [60,100,140] & & 64 & 0.01 & 120 & [40,80,100] \\
SS-v2 &    32&  0.0001 & 45  & [25, 35, 40] & & 64&  0.01 & 40 & [20,30] \\
EK-100 &   32&  0.0025 & 30  & [20, 25] & & 32 &  0.0025 & 30 & [20, 25] \\
K-400   & -& -& -&- & & 64&  0.01 & 40 & [10,20,30] \\
%\multicolumn{5}{c}{Linear Evaluation}\\
%\hdashline 
%UCF-101 & 64&  0.01 & 100 & 40,80 \\
%NTU-60 &  64&  0.01 & 120 & 40,80,100 \\
%Gym-99 &  64&  0.01 & 120 & 40,80,100 \\
%SS-v2 &   64&  0.01 & 40 & 20,30 \\
%EK-100 &  32 &  0.0025 & 30 & 20, 25 \\
%K-400  - & -& -& -& &   64&  0.01 & 40 & 10,20,30 \\
\bottomrule
\end{tabular}%
}
\label{tab:downstream_domain_training}
\end{table}

\subsection{Downstream Samples}
In \cref{sec:factor_2} we measure how sensitive current video self-supervised models are to the amount of downstream samples. We do this by varying the size of the training data starting from 1000 examples and doubling it until we reach the full train set. We use the same data splits as in the downstream domain experiments, explained in \cref{app:domain-shift-expt}, and sample a subset of video clips from the respective train sets. We use the same random subset across the different models % NumPy seeding to sample same training samples across different models 
to make the comparison fair. For each dataset, we use same training and testing procedure as the downstream domain experiments, explained in   \cref{app:domain-shift-expt} and \cref{tab:downstream_domain_training}. 

%measuring performance on different subsets, defined in the FineGym dataset~\cite{Gym-99-arxiv}, which have increasing semantic similarity. We provide the details of Gym 99, Gym 288 and the four different subsets we use of Gym 99 here:

\subsection{Downstream Actions}
%\section{FineGym Subsets}
In \cref{sec:factor_3} we measure how benchmark-sensitive current video self-supervised models are to downstream actions. We do so by measuring performance on different subsets, defined in the FineGym dataset~\cite{Gym-99-arxiv}, which have increasing semantic similarity. We provide the details of Gym-99, Gym-288 and the four different subsets we use of Gym-99 below: %and \cref{tab:finegym-description}:

\noindent\textbf{Gym-99} consists of 29k video clips of 99 different actions across the four different gymnastic events in FineGym: Vault, Floor Exercise, Balance Beam and Uneven Bars. This is a relatively balanced subset of the full FineGym dataset with all actions having more than 80 occurrences. There are a total 20.5k training videos and 8.5k testing videos.

\noindent\textbf{Vault} is a subset of Gym 99 containing 1.5k videos of the 6 actions from the Vault event. %as listed in \cref{tab:vault_actions}. 
The training split contains 1.0k examples and the testing split contains 0.5k examples.

\noindent\textbf{Floor} contains actions in the Floor Exercise event from Gym-99. It consists of 7.5k instances of over 35 actions with a split of 5.3k for training and 2.2k for testing. %The actions are listed in \cref{tab:floor_actions}.

\noindent\textbf{FX-S1} is a subset of actions of leaps, jumps and hops from the Floor event in Gym-99. This subset of 11 actions %is shown in \cref{tab:floor_actions} and 
contains a total of 2.6k video clips with 1.9k for training and 0.7k for testing.

\noindent\textbf{UB-S1} contains 5k videos of 15 actions from the Uneven Bars event with a split of 3.5k for training and 1.5k for testing. The actions consist of different types of circles around the bars. %and are listed in \cref{tab:uneven_bars}.

\noindent\textbf{Gym-288} is a long-tailed version of Gym 99 containing 32k videos with 22.6K training and 9.6K testing samples. It adds 189 infrequent classes to the 99 classes in Gym 99, where actions can have as little as 1 or 2 instances in training. This results in a total of 288 action classes from the four different gymnastic events. 

We follow the same training and evaluation procedure as that for finetuning Gym-99 in downstream domain training. In particular, for training we sample a random clip from each video of 32 frames with standard augmentations \ie a random multi-scale crop of size 112x112, random horizontal flipping and color jitter. Each model is trained with the Adam optimizer using a learning rate of 0.0001 and multi-step scheduler with $\gamma {=} 0.1$ at epochs [60, 100, 140] for 160 epochs. For inference, we use 10 linearly spaced clips of 32 frames each. For each frame we take a center crop which is resized to 112x112 pixels. To calculate the action class prediction of a video, we take the mean of the predictions from each clip. For each subset, we compute accuracy per action class and report the mean over all action classes as in the original dataset \cite{Gym-99-arxiv}.

\subsection{Downstream Tasks}
In \cref{sec:factor_4} we investigate how sensitive self-supervised methods are to the downstream task and whether they generalize beyond action recognition. We provide details of the experimental setup used for each task below.
% The hyperparameters used are summarized in \cref{tab:tasks-hparams}.

\noindent\textbf{Spatio-temporal action detection}. The goal of this task is to predict the bounding box of an actor in a given video clip, both spatially and temporally, along with the action class. We use the UCF101-24 benchmark which is a subset of UCF-101 with bounding box annotations for 3,207 videos from 24 action classes. We follow the implementation of K{\"{o}}p{\"{u}}kl{\"{u}} \etal \cite{yowo} using only a 3D-CNN branch for spatio-temporal action detection. We initialize the 3D backbone with the pre-trained, self-supervised R(2+1D)-18 models. A clip size of 16 frames is sampled from the video as the input with standard data augmentations \ie horizontal flipping, random scaling and random spatial cropping. Each model is trained using the Adam optimizer with an initial learning rate of 1e-4, weight decay of 5e-4 and batch size 64, for a total of 12 epochs. The learning rate is decayed using a multi-step scheduler with $\gamma {=} 0.5$ at epochs [4,6,8,10]. For testing we also follow \cite{yowo} and report video-mAP over all the action classes.
    
 \noindent\textbf{Repetition counting}. The goal 
 of the this task is to estimate the number of times an action repeats in a video clip. We use the UCFRep benchmark proposed by Zhang \etal \cite{rep_counting}, which is a subset of UCF-101. The dataset consists of 526 videos with 3,506 repetition number annotations. From the annotated videos, 2M sequences of 32 frames and spatial size 112x112 are constructed which are  used as the input. %and  follow the same training and evaluation procedure  as in \cite{rep_counting}. 
    We use the implementation from the original benchmark \cite{rep_counting} with pre-trained R(2+1)D-18 models as the backbone networks. Each model is trained for 100 epochs with a batch size of 32 using the Adam optimizer with a fixed learning rate of 0.00005. For testing, we follow the protocol from \cite{rep_counting} and report mean counting error.
    
\noindent \textbf{Arrow-of-time}. The goal of this task is to predict the direction (forward of backward) of the video. We closely follow the setup used by Ghodrati \etal \cite{arrow_of_time}. The full UCF-101 dataset is used with two versions of each video, one normal and one reversed. During training, for each video, we sample 8 frames linearly with a random offset, with batch size of 12 and 112x112 center crops, number of epochs 10, learning rate of $1e^{-5}$. We do not use any augmentations or learning rate schedulers. During testing, we sample 8 frames linearly. We report top-1 binary classification accuracy.

\noindent\textbf{Multi-label classification on Charades}. Charades \cite{charades-sigurdsson:hal-01418216} is made up of videos of people recording everyday activities at their homes.  %It differs in visual appearance and has longer actions with annotations for multi-label action classification task.  
Videos in Charades are longer than the other datasets we use and the goal is to recognize multiple different actions in each video. A per-class sigmoid output is used for multi-class prediction.
We use the implementation of Feichtenhofer \etal \cite{large-scale-feichtenhofer2021large}\footnote{\href{https://github.com/facebookresearch/SlowFast}{https://github.com/facebookresearch/SlowFast}} with the R(2+1)D-18 backbone. 
During training, we use 32 frames with a sampling rate of 8. Since this task requires longer temporal context, we observe that using more frames with higher sampling rate is beneficial. We use a spatial crop of 112x112 and augmentations such as random short-side scaling, random spatial crop and horizontal flip. We train for 57 epochs in total with a batch size of 16 and a learning rate of 0.0375 with multi-step scheduler with $\gamma = 0.1$ at epochs [41, 49]. During testing, following \cite{large-scale-feichtenhofer2021large}, we spatio-temporally max-pool predictions over 10 clips for a single video. We report mean average precision (mAP) across classes.

\noindent\textbf{Action detection on AVA.} AVA \cite{AVA-Gu_2018_CVPR} consists of clips extracted from films. We use version v2.2 with bounding box annotations for spatio-temporal  action detection of temporally fine-grained action classes.
% Like Kinetics-400, it has high variability in visual appearance. 
The goal of this task is to detect and predict action classes from proposals generated by off-the-shelf person detectors. We again use the implementation of \cite{large-scale-feichtenhofer2021large} with the R(2+1)D-18 backbone. During training, we use 32 frames with a sampling rate of 2. We use spatial crop of 112x112 and augmentations such as random short-side scaling, random spatial crop, horizontal flip. We train for 20 epochs with learning rate of 0.1 with multi-step scheduler with $\gamma = 0.1$ at epochs [10, 15] and a batch size of 32. During testing, following \cite{large-scale-feichtenhofer2021large}, we use a single clip at the center of the video with 8 frames and sampling rate of 8. We report mean average precision (mAP) across the classes.

\section{Correlations of Downstream Performance}
\label{correlattion_plots}

\begin{figure}[htb!]
\captionsetup{font=footnotesize,skip=1mm}
    \centering
    \includegraphics[width=\linewidth]{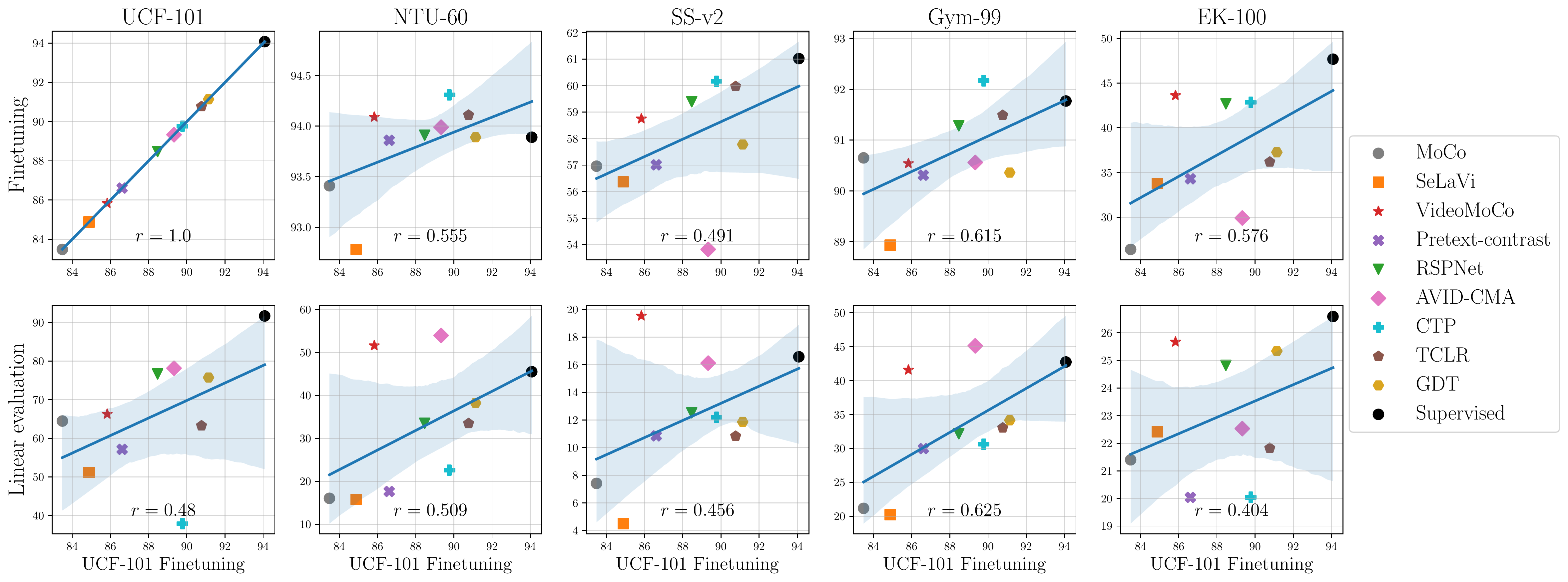}
    \caption{\textbf{Downstream domain against UCF-101 finetuning.} 
    We plot the corelations between finetuning performance of video pre-training methods on UCF-101  and performances on finetuning and linear-evaluation on all downstream datasets.}
    \label{fig:corr_on_ucf}
\end{figure}

\begin{figure}[h]
\captionsetup{font=footnotesize,skip=1mm}
    \centering
    \includegraphics[width=0.95\linewidth]{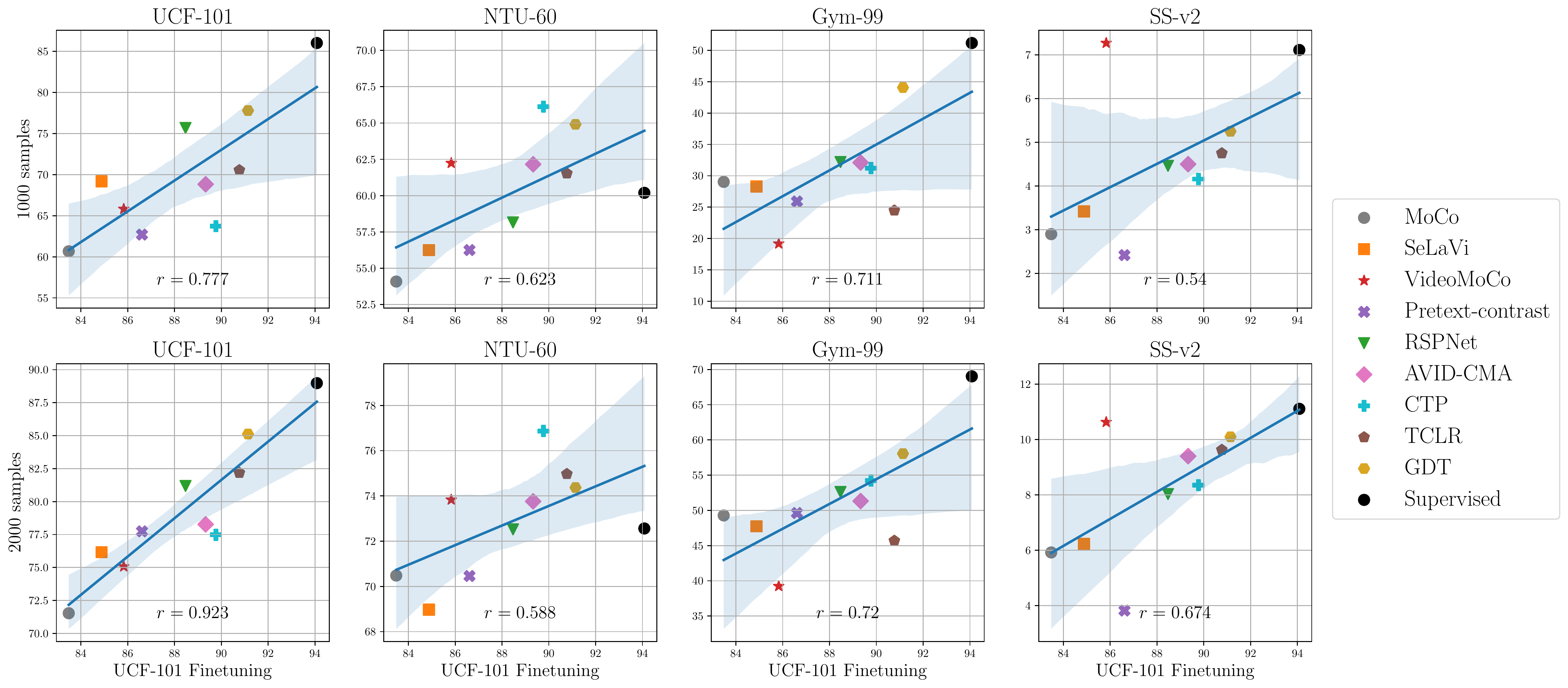}
    \caption{\textbf{Downstream samples against UCF-101 finetuning.} 
    For the low data setting (1000-2000 samples), we plot the correlations of performance of video pre-training methods  against that for UCF-101 finetuning.}
    \label{fig:samples-corr_on_ucf}
\end{figure}

\begin{figure}[htb!]
\captionsetup{font=footnotesize,skip=1mm}
    \centering
    \includegraphics[width=\linewidth]{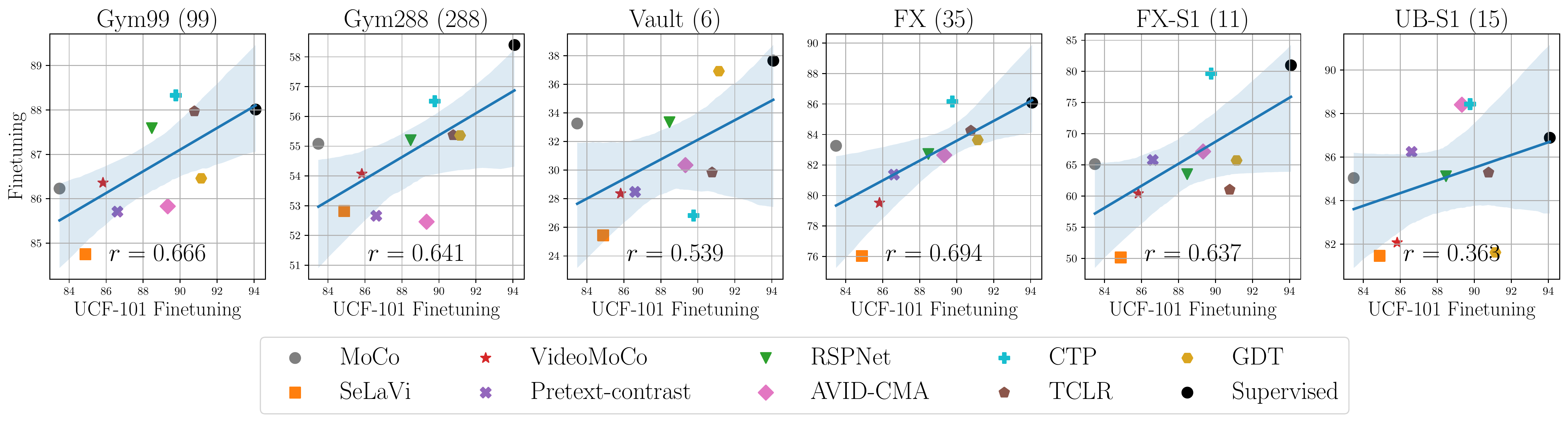}
    \caption{\textbf{Downstream actions against UCF-101 finetuning.}
     We plot the corelations of performances of video pre-training methods between  UCF-101 finetuning and  FineGym subsets.}
    \label{fig:actions-corr_on_ucf}
\end{figure}

\begin{figure}[htb!]
\captionsetup{font=footnotesize,skip=1mm}
    \centering
    \includegraphics[width=\linewidth]{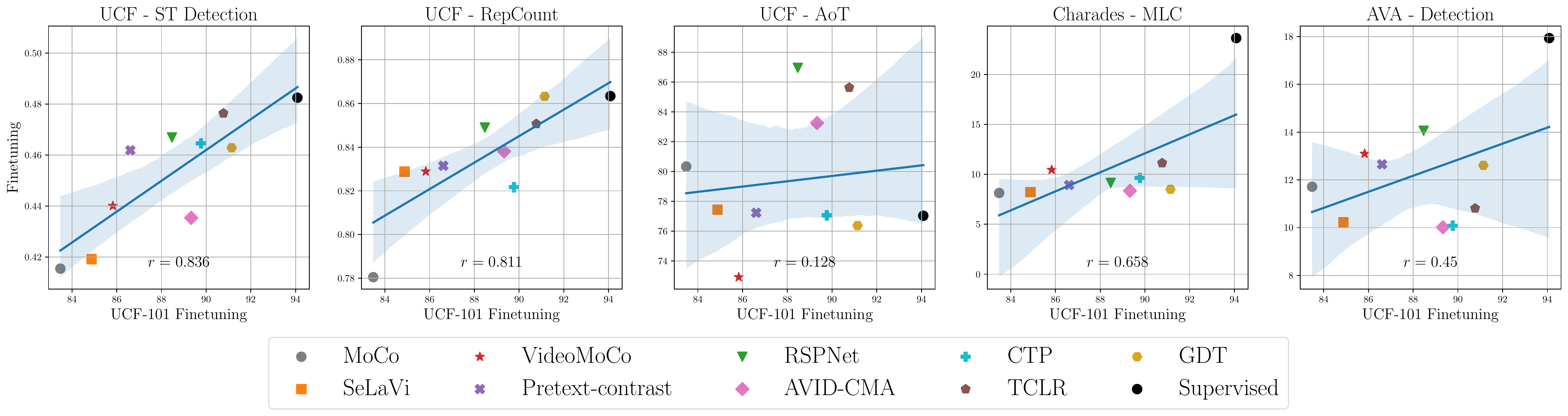}
    \caption{\textbf{Downstream tasks against UCF-101 finetuning.}
     We plot the corelations between performance on UCF-101 finetuning  and other downstream tasks for the video pre-training methods.}
    \label{fig:tasks-corr_on_ucf}
\end{figure}

\begin{figure}[htb!]
\captionsetup{font=footnotesize,skip=1mm}
    \centering
    \includegraphics[width=\linewidth]{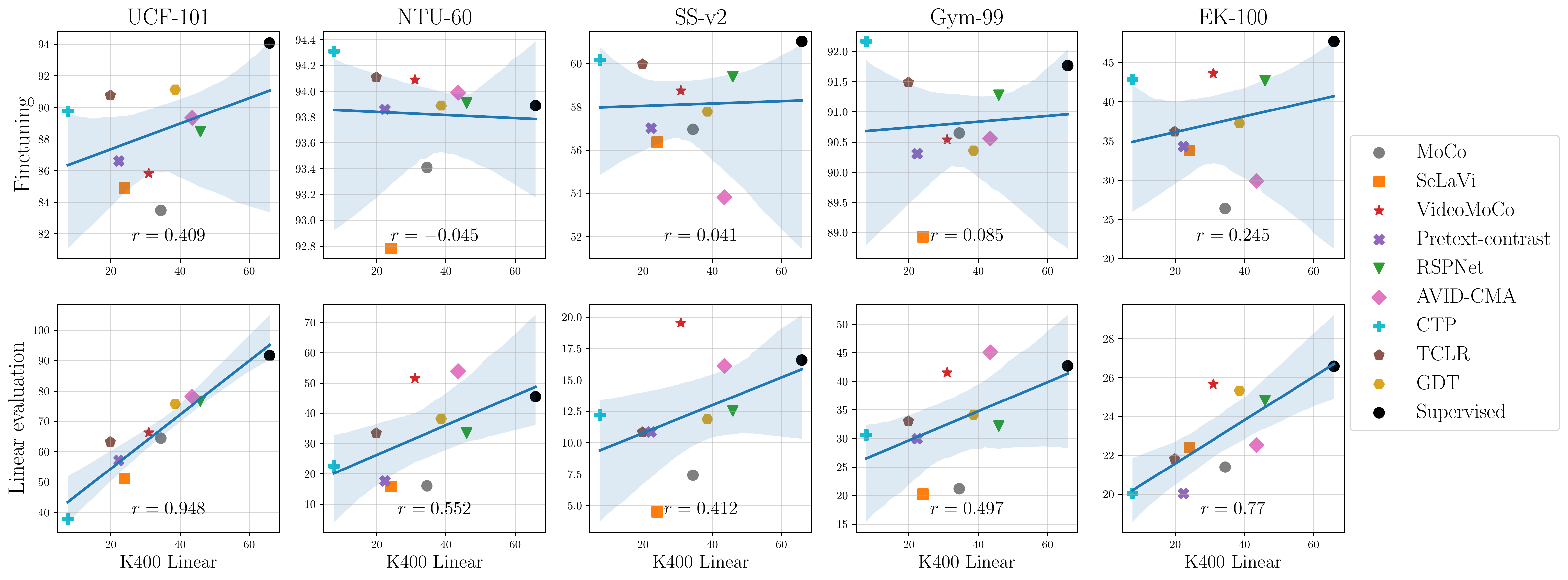}
    \caption{\textbf{Downstream domain against Kinetics-400 linear evaluation.} 
    We plot the corelations between finetuning performance of video pre-training methods on Kinetics-400 linear-evaluation  and performances on finetuning and linear-evaluation on all downstream datasets.}
    \label{fig:domains-corr_on_k400}
\end{figure}

\begin{figure}[htb!]
\captionsetup{font=footnotesize,skip=1mm}
    \centering
    \includegraphics[width=\linewidth]{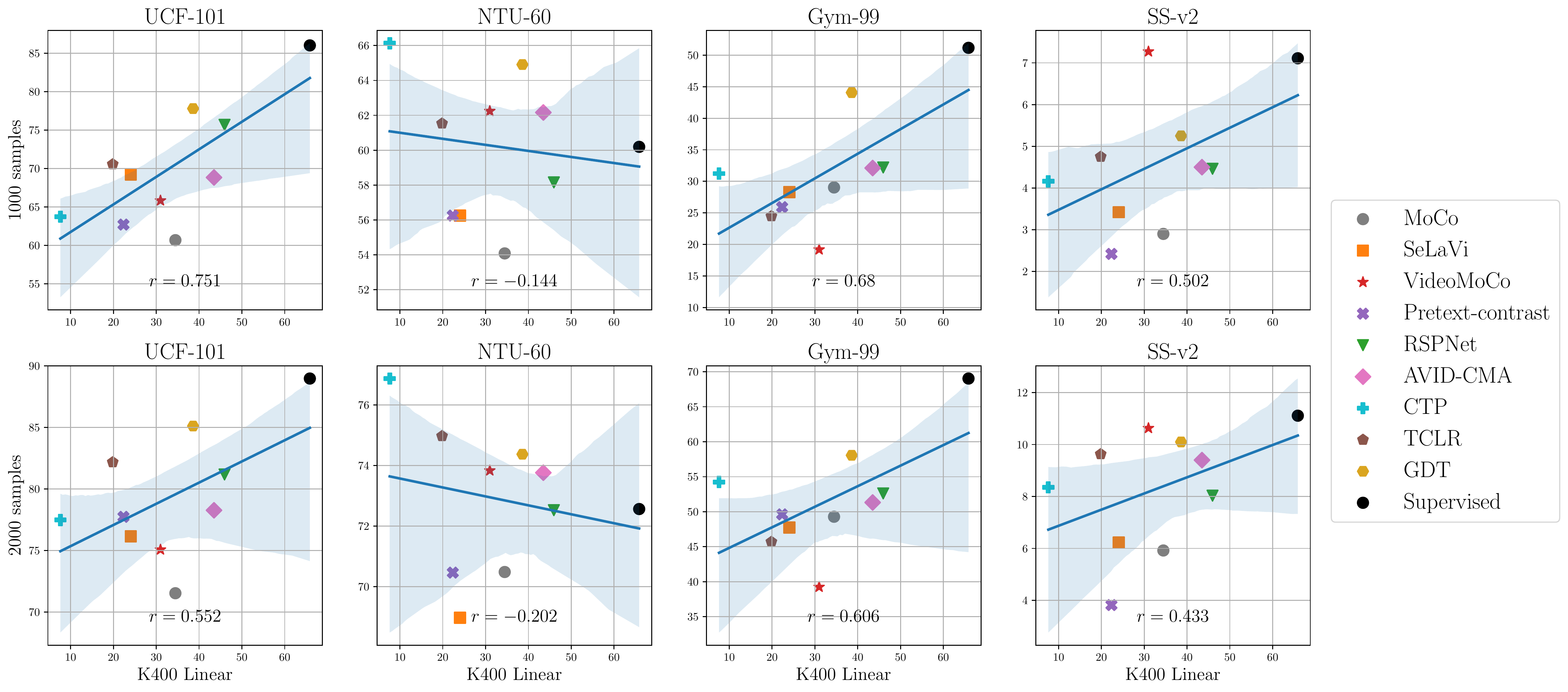}
    \caption{\textbf{Downstream samples against Kinetics-400 linear evaluation.} 
    For the low data setting (1000-2000 samples), we plot the correlations of performance of video pre-training methods  against that for Kinetics-400 linear-evaluation.}
    \label{fig:samples-corr_on_k400}
\end{figure}

\begin{figure}[htb!]
\captionsetup{font=footnotesize,skip=1mm}
    \centering
    \includegraphics[width=\linewidth]{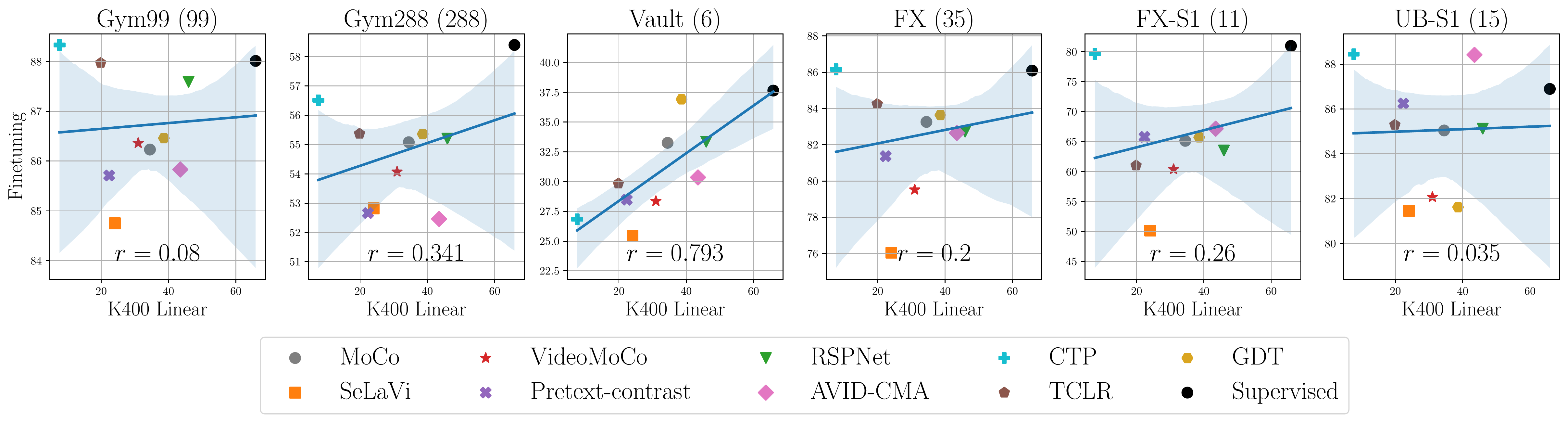}
    \caption{\textbf{Downstream actions against Kinetics-400 linear evaluation.} 
     We plot the corelations of performances of video pre-training methods between  Kinetics-400 linear-evaluation and  FineGym subsets.}
    \label{fig:actions-corr_on_k400}
\end{figure}

\begin{figure}[htb!]
\captionsetup{font=footnotesize,skip=1mm}
    \centering
    \includegraphics[width=\linewidth]{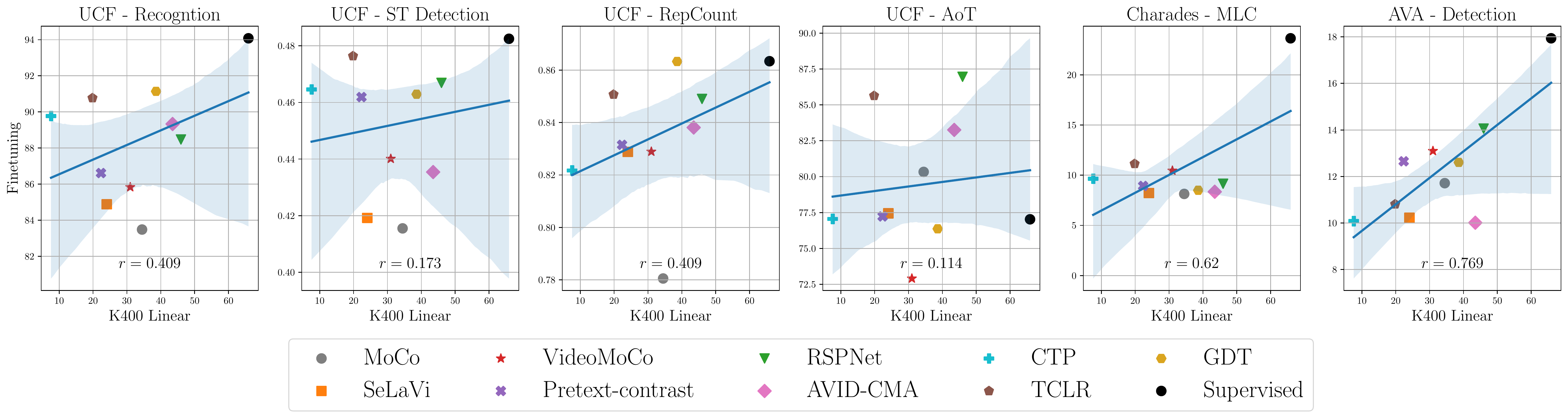}
    \caption{\textbf{Downstream tasks against Kinetics-400 linear evaluation.}
     We plot the corelations between performance on Kinetics-400 linear-evaluation  and other downstream tasks for the video pre-training methods.}
    \label{fig:tasks-corr_on_k400}
\end{figure}

%with the ranking of methods changing substantially across datasets and whether full finetuning or linear classification is used.
As observed from the results of \cref{sec:factor_1},  the performance for both UCF-101 finetuning and Kinetics-400 linear evaluation is not indicative of how well a self-supervised video model generalizes to different downstream domains, samples, actions and tasks.
Here, we  plot the performance of each pre-trained model for each downstream settings and show the correlation with UCF-101 finetuning and Kinetics-400 linear evaluation performances. The results are shown  in \cref{fig:corr_on_ucf,fig:samples-corr_on_ucf,fig:actions-corr_on_ucf,fig:tasks-corr_on_ucf,fig:domains-corr_on_k400,fig:samples-corr_on_k400,fig:actions-corr_on_k400,fig:tasks-corr_on_k400}. These plots further demonstrate that the correlations are overall low for each downstream factor  \ie domain, samples, actions and tasks, indicating that more thorough testing of video self-supervised methods is needed. 

% \begin{figure}
% \captionsetup{font=footnotesize,skip=1mm}
%     \centering
%     \includegraphics[width=\linewidth]{media/correlations_domain_shift_on_K400_v1.pdf}
%     \includegraphics[width=\linewidth]{media/correlations_action_granularity_on_K400-linear_v1.pdf}
%     \includegraphics[width=\linewidth]{media/correlations_task_shift_on_K400-linear_v1.pdf}
%     \caption{\textbf{Correlations between downstream task performance and  Kinetics-400 linear-classification accuracy.}}
%     \label{fig:corr_on_k400}
% \end{figure}

\clearpage
\section{Representation Similarity Matrices}
\label{similarity_features}
\begin{figure}
\captionsetup{font=footnotesize,skip=1mm}
    \includegraphics[width=\linewidth]{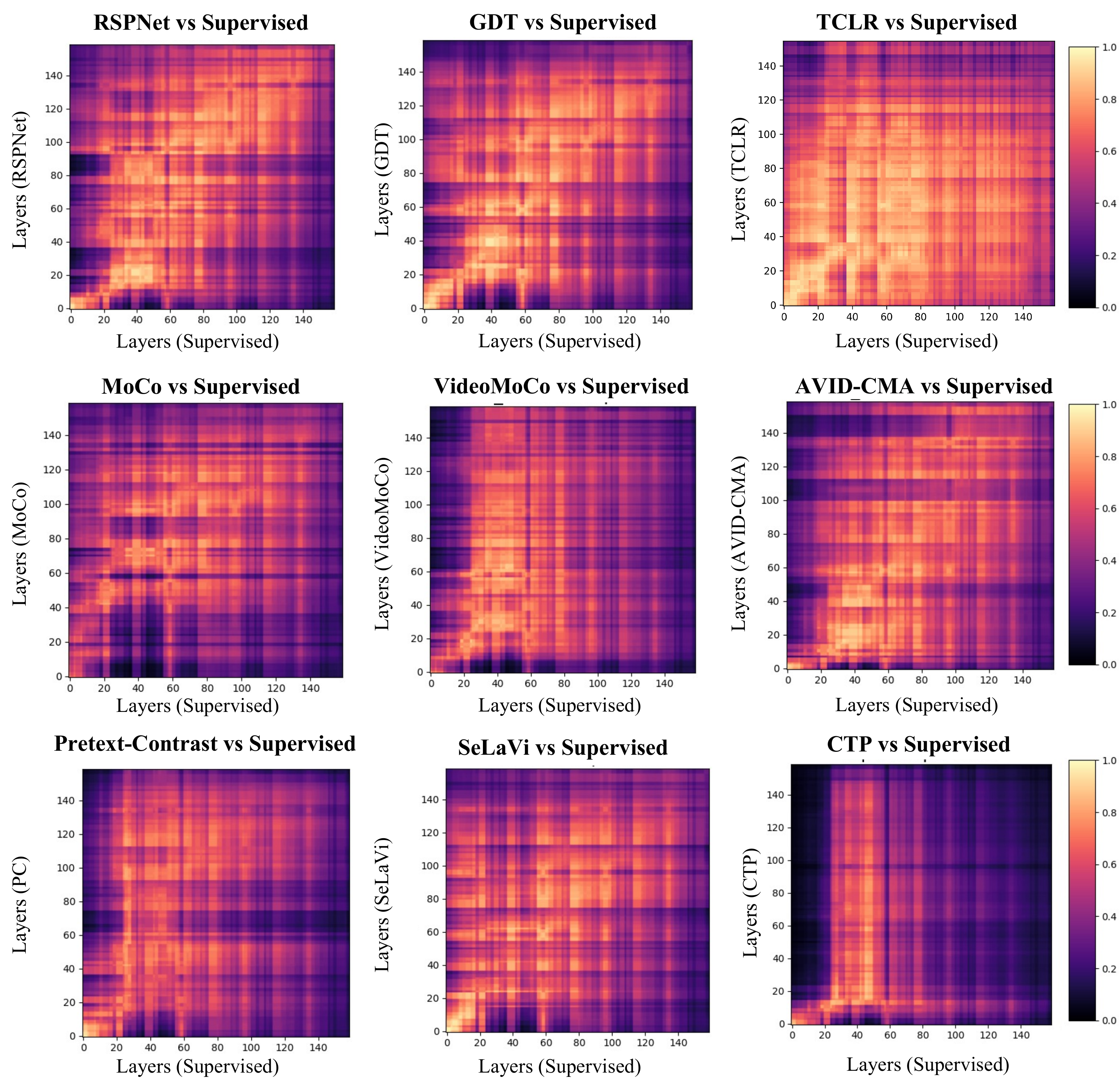}
     \caption{\textbf{Representation similarity} between features of self-supervised methods and supervised pre-training on Kinetics-400 validation set using centered kernel alignment. Features of contrastive methods are more closer to the features of supervised pretraining. %especially the ones which encourage temporal distinctivenss \ie RSP
     }
    \label{fig:cka_all}
\end{figure}

We plot the the feature similarity on Kinetics validation set using centered kernel alignment ~\cite{cka} between supervised pre-training and our evaluated self-supervised pre-training methods in \cref{fig:cka_all}. We showed a subset of these plots in \cref{features}, here we show the feature similarity for all the self-supervised models we used in our experiments.

\section{Downstream Dataset Attributes}
\label{sec:video-datasets}

% \pb{Details of how these datasets were created/collected/recorded, number of classes, number of video clips, etc.}

\begin{figure}
    \centering
    \includegraphics[width=\linewidth]{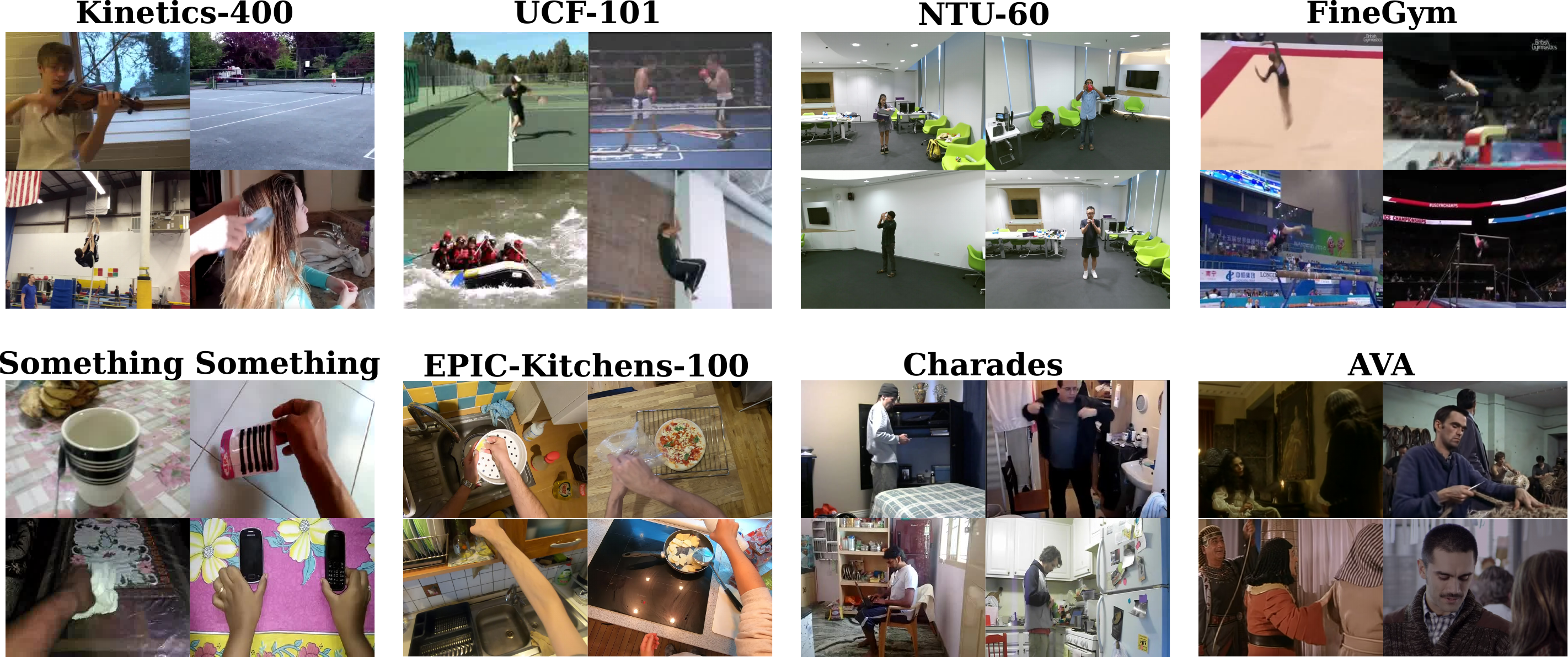}
    \caption{Example video frames from the Kinetics-400 pre-training dataset and the 7 different downstream datasets we consider. Note the differences in the capture setting and point-of-view across these datasets.}
    \label{domain_frames_appendix}
\end{figure}

\begin{figure}
    \centering
    \begin{tabular}{@{}c@{}c}
        \includegraphics[width=0.8\linewidth]{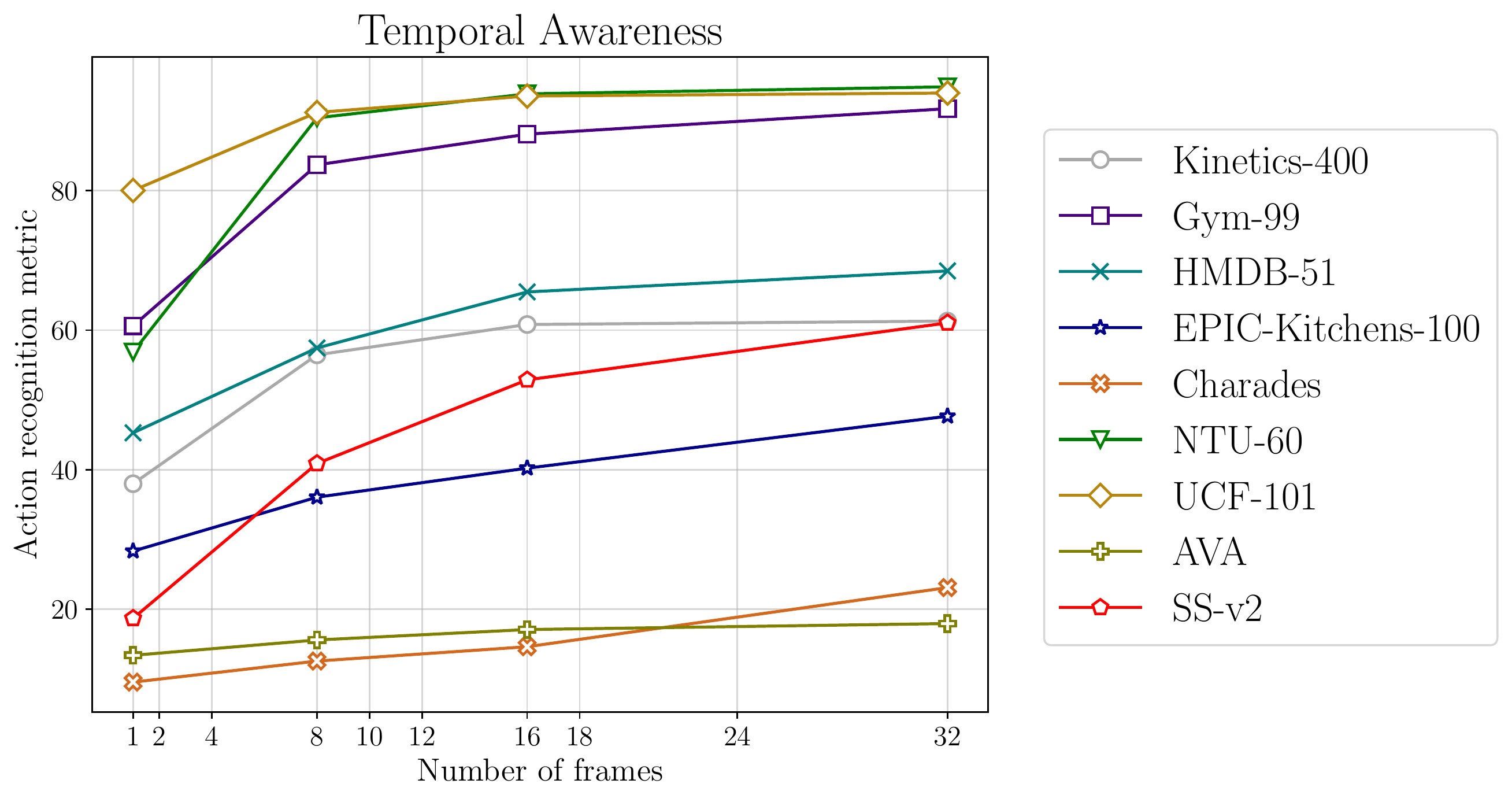} 
    \end{tabular}
    \caption{\small \textbf{Temporal awareness}. Illustrating the effect of temporal awareness (increasing temporal-context)  on the action recognition performance using a standard 3D-CNN for different action datasets.
    % See appendix for more details.
    }
    \label{fig:action_temporality-intradataset}
\end{figure}

\begin{figure}[t]
    \centering
    \begin{tabular}{@{}c@{}c}
        \includegraphics[width=0.95\linewidth]{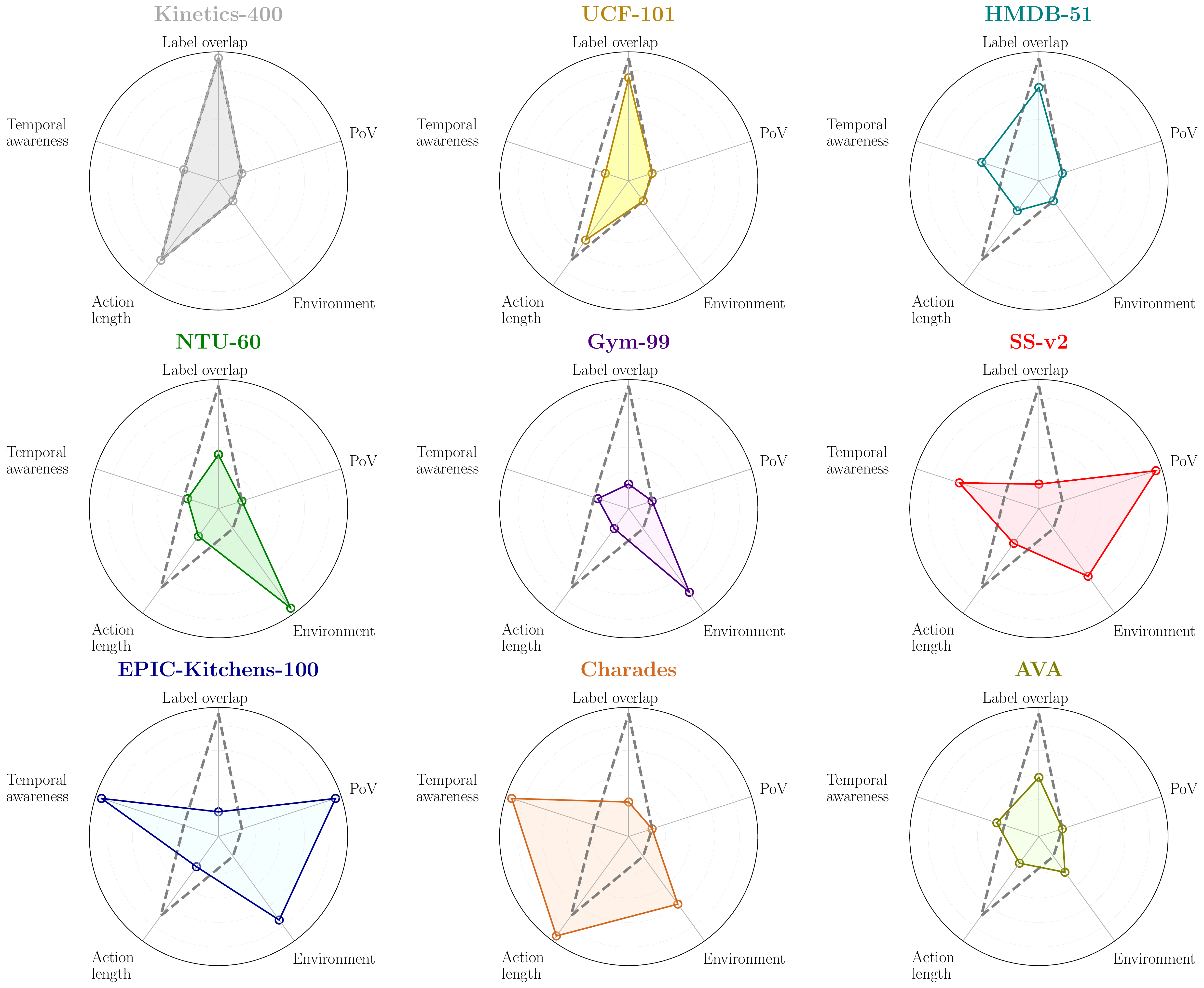} 
    \end{tabular}
    \caption{\small \textbf{Radar plots with details}. The radar plots contain details of the values along the axis for every attribute for the datasets we use in this study.}
    \label{fig:radar-detailed}
\end{figure}
We define several attributes in \cref{subsec:domain-shift} in order to characterize differences in domain between the downstream datasets and the Kinetics-400 pre-training dataset in \cref{fig:radar}. We provide detailed radar plots in \cref{fig:radar-detailed} with axes labeled with relevant values for each attribute. % the shift in domain of a target dataset with the source dataset, Kinetics-400. 
The attributes \textit{Point-of-view} and \textit{Environment} are defined qualitatively based on the contents of the target dataset. Examples of videos from each of the datasets are shown in \cref{domain_frames_appendix}. We can see that FineGym \cite{Gym-99-arxiv} consists of videos of Olympic gymnastic events. Thus, we label it as \textit{stadium} for environment and \textit{third-person} for point-of-view. On the radar plots, we order environment in descending order of variability contained in a given dataset. Kinetics-400 is placed near the origin as it has much higher variability than NTU-60 for example, which is captured in a controlled lab setting. \textit{Action length} is the average duration of the actions in each of the datasets. 

We quantify \textit{temporal awareness} as the minimum number of frames (temporal context) required to best recognize the action. We do this by finetuning R(2+1)D with weights initialized from supervised pre-training on Kinetics-400 and we denote temporal awareness ($\tau$) as:
\begin{equation}
    \tau = \arg\min_{t \in\{1, 2, ..., N\}}\left[ \left(100 \times \frac{f_{t+1} - f_{t}}{f_{t}}\right) < \alpha \right]
\end{equation}
where $\alpha$ is chosen to be $1$. This means $\tau$ indicates the number of frames after which relative improvement in performance is lesser than $\alpha$, \ie when the performance has plateaued. \cref{fig:action_temporality-intradataset} shows the top-1 action recognition performance against increasing number of frames for each of our downstream datasets. We use bilinear interpolation to estimate performance at given number of frames beyond those that we experiments with. For example, using our method to compute temporal awareness, the performance for UCF-101 plateaus at 7 frames while that for EK-100 plateaus at 32 frames indicating that EK-100 needs much larger temporal context for recognition while UCF-101 may suffice with a shorter temporal context.

\textit{Label overlap} is the amount of actions which are present in both the downstream dataset and the pretraining dataset (Kinetics-400). We quantify this by matching identical actions as well as manually checking for reworded versions of the same action class. % by searching through the most similar actions in Kinetics-400 for every action in the target dataset. 
For example, ``head massage'' in UCF-101 has a corresponding action ``massaging person's head'' in Kinetics-400. In NTU-60 action class ``brushing teeth'' has a matching action ``brushing teeth'' in Kinetics-400. % and so on.
% We provide the entire lists of matching actions in \hd{Table XYZ}.

%\clearpage
\section{Standard deviations for proposed SEVERE-benchmark}
\label{sec:std_dev}
In this section, we show the standard deviations of each pretrained method for all downstream setups in our proposed benchmark. The results are obtained  over 3 runs initialized with different random seeds. It is clear from Table~\ref{proposed-benchmarks-mean} that results are consistent over multiple runs with small $std$ deviations. Thus our observations and  conclusions are not impacted across multiple runs. Moreover, future works can refer to Table~\ref{proposed-benchmarks-mean} for reproduciblity.

%\begin{adjustbox}{angle=90}
\begin{table}[h]
\captionsetup{font=small,skip=2mm}
         \caption[]{\textbf{Standard deviations for proposed SEVERE-benchmark}. We compute the $std$ of each method for each downstream setup  over 3 runs initialized with random seeds.
    }
    \centering
    \midsepremove
    \resizebox{\linewidth}{!}{\begin{tabular}
    {
    l
    \C{77.3}{93.9}
    \C{47.7}{68.5}c
    \C{52.0}{60.8}\C{89.9}{92.1}@{\hskip 2mm}c
    \C{38.3}{86.6}\C{22.7}{51.3}@{\hskip 2mm}c
    \C{46.6}{79.1}\C{80.9}{88.8}@{\hskip 2mm}c
    \CR{0.123}{0.217}\C{7.9}{23.5}
    }
    % {lc cc ccccccc}
    \toprule
    \addlinespace[0.1cm]
     & \multicolumn{2}{Sc}{\textbf{Existing}} & & \multicolumn{11}{Sc}{\textbf{SEVERE-benchmark}} \\
    \addlinespace[0.04cm]
    \cmidrule{2-3} \cmidrule{5-15}
      \addlinespace[0.1cm]
         %& \multicolumn{1}{c}{} & & \multicolumn{2}{Sc}{Domains} &  & \multicolumn{2}{Sc}{Samples} & & \multicolumn{2}{Sc}{Actions}& & \multicolumn{2}{Sc}{Tasks}\\
         \multicolumn{1}{l}{\textbf{Pre-training}} & \multicolumn{2}{c}{} & & \multicolumn{2}{Sc}{Domains} &  & \multicolumn{2}{Sc}{Samples} & & \multicolumn{2}{Sc}{Actions}& & \multicolumn{2}{Sc}{Tasks}\\
    %   \cmidrule{4-11}
      \cmidrule(lr){5-6} \cmidrule(lr){8-9} \cmidrule(lr){11-12} \cmidrule(lr){14-15}
      \addlinespace[0.1cm]
         & \multicolumn{1}{c}{UCF101} & \multicolumn{1}{c}{HMDB51} & & \multicolumn{1}{c}{SS-v2} & \multicolumn{1}{c}{Gym-99} & & \multicolumn{1}{c}{UCF ($10^{3}$)} & \multicolumn{1}{c}{Gym-99 ($10^{3}$)} & & \multicolumn{1}{c}{FX-S1} & \multicolumn{1}{c}{UB-S1}& & \multicolumn{1}{c}{UCF-RC} & \multicolumn{1}{c}{Charades-MLC}\\
         \midrule
         
           \addlinespace[0.01cm]
         None                 & 77.3 {\pm 0.9} &47.7 {\pm 1.6}  && 57.1 {\pm 1.3} & 89.8 {\pm 0.1} && 38.3 {\pm 1.4} & 22.7 {\pm 3.5} && 46.6 {\pm 1.8} & 82.3 {\pm 2.1} && 0.217 {\pm 0.01} & 7.9 {\pm 0.1} \\
           \addlinespace[0.01cm]
         \midrule
        \addlinespace[0.01cm]
         MoCo                 & 83.3 {\pm 0.3} & 53.6 {\pm 0.2}  && 57.1 {\pm 0.1}   & 90.7 {\pm 0.2}  && 60.4 {\pm 1.0}  & 30.9 {\pm 1.0}  && 65.0 {\pm 1.2}  & 84.5  {\pm 0.4} && 0.208 {\pm 0.01}  & 8.3  {\pm 0.1}  \\
         VideoMoCo            & 84.9 {\pm 0.5} & 58.0 {\pm 1.0}  && 59.0 {\pm 0.1}  & 90.3 {\pm 0.3}  && 65.4 {\pm 1.2}  & 20.6 {\pm 0.8}  && 57.3  {\pm 2.9} & 83.9 {\pm 1.6}  && 0.185 {\pm 0.00}  & 10.5  {\pm 0.1} \\
         SeLaVi               & 85.2 {\pm 0.3} & 54.2 {\pm 0.3}   && 56.2 {\pm 0.1}  & 88.9 {\pm 0.1}  && 69.0 {\pm 1.9}  & 30.2 {\pm 0.9}  && 51.3 {\pm 1.0}  & 80.9 {\pm 1.6}   && 0.162 {\pm 0.01}  & 8.4  {\pm 0.1}  \\
         Pretext-Contrast     & 87.7 {\pm 0.6} & 58.4 {\pm 0.6}  && 56.9 {\pm 0.2}  & 90.5  {\pm 0.1} && 64.6 {\pm 2.3}  & 27.5 {\pm 1.6}  && 66.1 {\pm 0.3}  & 86.1  {\pm 0.8} && 0.164 {\pm 0.01}  & 8.9   {\pm 0.1} \\
         RSPNet               & 88.7 {\pm 0.1} & 59.2 {\pm 0.7}  && 59.0 {\pm 0.3}  & 91.1  {\pm 0.0} && 74.7 {\pm 0.6}  & 32.2 {\pm 1.5}  && 65.4  {\pm 1.7} & 83.6 {\pm 1.3}  && 0.145 {\pm 0.01}  & 9.0  {\pm 0.3}  \\
         AVID-CMA             & 88.8 {\pm 0.3} & 58.7 {\pm 1.2}  && 52.0 {\pm 0.6}  & 90.4 {\pm 0.4}  && 68.2 {\pm 0.5}  & 33.4 {\pm 0.8}  && 68.0 {\pm 0.9}  & 87.3 {\pm 1.0}  && 0.148 {\pm 0.01}  & 8.2  {\pm 0.2}  \\
         CtP                  & 90.1 {\pm 0.1} & 63.2 {\pm 0.5}  && 59.6 {\pm 0.4}  & 92.0 {\pm 0.1}  && 61.0 {\pm 1.5}  & 32.9 {\pm 1.9}  && 79.1 {\pm 0.5}  & 88.8  {\pm 0.5} && 0.178 {\pm 0.01}  & 9.6  {\pm 0.1}  \\
         TCLR                 & 90.8 {\pm 0.2} & 60.6 {\pm 0.9}  && 59.8  {\pm 0.0} & 91.6 {\pm 0.0}  && 72.6 {\pm 1.9}  & 26.3 {\pm 1.0}  && 60.7 {\pm 0.7}  & 84.7  {\pm 1.1} && 0.142 {\pm 0.01}  & 12.2 {\pm 0.3}  \\
         GDT                  & 91.3 {\pm 0.3} & 64.8 {\pm 1.0}  && 58.0  {\pm 0.3} & 90.5 {\pm 0.1}  && 78.4 {\pm 0.2}  & 45.6 {\pm 0.6}  && 66.0 {\pm 0.3}  & 83.4  {\pm 1.6} && 0.123 {\pm 0.01}  & 8.5  {\pm 0.1}  \\
        \addlinespace[0.01cm]
         \midrule
        \addlinespace[0.01cm]
         Supervised           & 93.9 {\pm 0.2} & 68.5 {\pm 0.4}  && 60.8 {\pm 0.1}  & 92.1 {\pm 0.1}  && 86.6 {\pm 0.6}  & 51.3 {\pm 0.1}  && 79.0  {\pm 2.0} & 87.1  {\pm 0.2} && 0.132  {\pm 0.01} & 23.5  {\pm 0.1} \\
        \addlinespace[0.01cm]
         \bottomrule
    \end{tabular}
    }

    \label{proposed-benchmarks-mean}
\end{table}
%\end{adjustbox}

\section{HMDB51 Results}
\label{hmdb}
For completion we also show the performance of each pretraining method on HMDB51\cite{HMDB-51-ICCV} dataset in Table~\ref{proposed-benchmarks-mean}. HMDB51 is used in  current standard benchmarking along with UCF101\cite{UCF-101-arxiv}. It is clear from the table that the performances on both datasets are highly correlated to each other and less correlated to other downstream setups. This again highlights the importance of evaluating video self-supervised methods beyond current benchmarks and use a setup which evaluates the generalizability of current models, such as the SEVERE-benchmark.

%\clearpage
\section{Linear Evaluation for Downstream Samples}
\label{sec:lin_eval_samples}
\begin{figure}[h]
\captionsetup{font=small,skip=2mm}
    \centering
    \includegraphics[width=\linewidth]{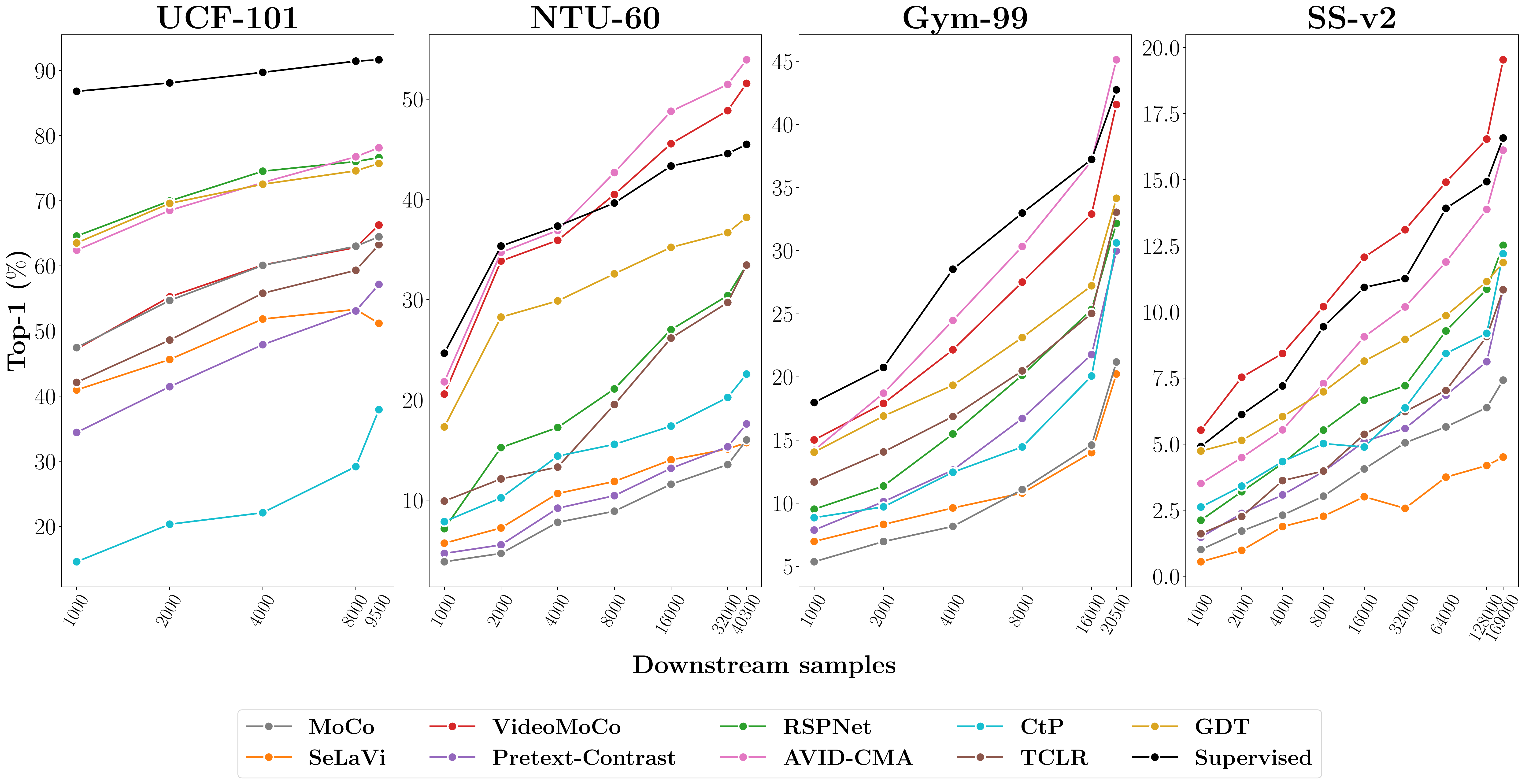}
    \caption{\textbf{Linear evaluation for Downstream Samples.} Comparison of video self-supervised learning methods using varying number of samples on linear evaluation for four downstream datasets.
    Rank changes are less significant  with increasing sample size.
    }
    \label{fig:training-data-size-linear}
\end{figure}

In \cref{sec:factor_2} we evaluated our pre-trained models  with varying amounts of downstream samples for finetuning. In this section we provide the results for the same experiment but using linear-evaluation instead of finetuning. The results are shown in \cref{fig:training-data-size-linear}. We observe that rank changes are not significant between different sample sizes as they are for full finetuning., However similar to finetuning, supervised pretraining is dominant for low data setting as shown by the performance on  NTU-60 and GYM-99 with 1000-4000 examples.
%\hd{results are in google sheet, piyush to plot and add here}

\section{Verb vs. Noun in Downstream Action Recognition}
\label{sec:epic_nouns}
EPIC-Kitchens-100 \cite{EPIC-100-arxiv} consists of noun and verb annotations for each video. An action is defined as a tuple of these. In the main paper, we report verb recognition performance across all experiments. In \cref{additional-expts-epic} we compare the performance on verb recognition to the performance on noun and action recognition.
In general, performance is lower for noun and action recognition in comparison to verb recognition. This is likely due to the R(2+1)D-18 backbone being insufficient to model the complex actions found in EPIC-Kitchens-100. Interestingly, good performance on  verb recognition is not a good indication that the model will perform well at noun or action recognition. Notably, some methods such as VideoMoCo and CtP perform well at verb recognition but struggle on noun recognition. RSPNet seems to perform reasonably well for both verb and noun recognition.

\begin{table}[t]
\centering
    \midsepremove
\captionsetup{font=small,skip=2mm}
\caption[]{\textbf{Ablation on Verb and Noun Recognition.} On EPIC-Kitchens-100, we show results for noun, verb and action recognition. Colors denote relative rankings across methods for each dataset, ranging from \textcolor{lowcolor}{low} \begin{tikzpicture}%
      \pgfplotscolorbardrawstandalone[%
        colormap name=PiYG,%
        colorbar horizontal,%
        colorbar style={%
          height=0.18cm,%
          width=2cm,%
          hide axis,%
        }%
      ]%
    \end{tikzpicture} \textcolor{highcolor}{high}. Most pre-training methods struggle on noun and action recognition and have little correlation with verb recognition. %\pb{What is the conclusion? Can add two more experiments to the same table, if any.}
     }
\setlength{\tabcolsep}{3mm}
\resizebox{0.4\textwidth}{!}{%
\begin{tabular}{l\C{25.7}{47.7}\C{6.9}{24.5}\C{1.8}{16.0}}%c\C{0.4}{2.3}\C{0.03}{1.82}\C{0.00}{0.69}}
\toprule
\addlinespace[0.07cm]
 & \multicolumn{3}{c}{\textbf{EPIC-Kitchens-100}} \\ 
\addlinespace[0.04cm]
% \cmidrule{2-5}\cmidrule{7-8} 
\cmidrule{2-4}
\addlinespace[0.04cm]
\addlinespace[0.04cm]
%& \multicolumn{3}{c}{Action Recognition} \\%&& \multicolumn{3}{c}{Action Detection}\\
%\cmidrule{2-4} \cmidrule{6-8}
\addlinespace[0.02cm]
\multicolumn{1}{l}{\textbf{Pre-training}} & \multicolumn{1}{c}{Verb} & \multicolumn{1}{c}{Noun} & \multicolumn{1}{c}{Action} \\%&& \multicolumn{1}{c}{Verb} & \multicolumn{1}{c}{Noun}  &  \multicolumn{1}{c}{Action} \\ 
%\multicolumn{1}{c}{} & \multicolumn{1}{c}{Recognition} & \multicolumn{1}{c}{Recognition} & \multicolumn{1}{c}{Recognition} &\multicolumn{1}{c}{Detection} & \multicolumn{1}{c}{Detection}  &  \multicolumn{1}{c}{Detection} \\ 
\addlinespace[0.02cm]
\midrule
\addlinespace[0.01cm]
% \arrayrulecolor{lightgray}\specialrule{0.05pt}{0.15pc}{0.15pc}
None & 25.7 & 6.9 & 1.8 \\%&& 0.5 & 0.08 & 0.02  \\
\addlinespace[0.01cm]\midrule        \addlinespace[0.01cm]
MoCo & 26.4 & 13.9 & 6.9 \\%&& 1.1 & 0.55 & 0.18 \\
SeLaVi & 33.8 & 12.1 & 5.9 \\%&& 1.3 & 0.46 & 0.15 \\
VideoMoCo & 43.6 &	15.1 & 9.4 \\%&& 1.5 & 0.52 & 0.15 \\
Pretext-contrast & 34.3 & 11.4 & 5.6 \\%&& 1.1 & 0.44 & 0.14 \\
RSPNet & 42.7 &  18.7 & 11.7 \\%&& 1.7 & 0.89 & 0.38 \\
AVID-CMA & 29.9 & 8.7 & 3.6 \\%&& 0.7 & 0.20 & 0.05 \\
CtP & 42.8 & 12.0 & 7.8 \\%&& 1.3 & 0.29 & 0.09 \\
TCLR & 36.2 & 11.7 & 5.8 \\%&& 1.1 & 0.36 & 0.10 \\
GDT & 37.3 & 15.5 & 8.4 \\%&& 0.4 & 0.03 &  0.00 \\
\addlinespace[0.01cm]\midrule         \addlinespace[0.01cm]
Supervised & 47.7 & 24.5 & 16.0 \\%&& 2.3 & 1.82 & 0.69 \\
\addlinespace[0.01cm]
\bottomrule
\end{tabular}%
}
\label{additional-expts-epic}
\end{table}

\end{document}